%% file: thesis.tex
\documentclass{thesisclass}

% ...
\usepackage{multirow}
\usepackage{xspace}
\usepackage{times}
\usepackage{epsfig}
\usepackage{graphicx}
\usepackage{amsmath}
\usepackage{amssymb}
\usepackage{array}
\usepackage{xcolor}
\usepackage{ulem}
\usepackage[font=small,labelfont=bf,justification=justified,format=plain]{caption}
\usepackage{icomma}

\hypersetup{
 pdfauthor={Andreas Eberle},
 pdftitle={Pose-Driven Deep Models for Person Re-Identification},
 pdfsubject={Not set},
 pdfkeywords={Not set}
}

\newcommand{\dtauthor}{Andreas Eberle}
\newcommand{\mytitle}{Pose-Driven Deep Models for Person Re-Identification}

\newcommand{\reviewerone}{Prof. Dr.-Ing. Rainer Stiefelhagen}
\newcommand{\reviewertwo}{Prof. Dr.-Ing. Jürgen Beyerer}
\newcommand{\advisor}{Dr.-Ing. Saquib Sarfraz\\
& & Dipl.-Inform. Arne Schumann}

\newcommand{\timestart}{31. August 2017}
\newcommand{\timeend}{28. February 2018}

\newcommand{\dttit}{Pose-Driven Deep Models for Person Re-Identification}
\newcommand{\dtdateother}{Karlsruhe, 28. February 2018}
\newcommand{\dtaddress}{Andreas Eberle\\
                        email@andreas-eberle.com}

\newcommand{\kindg}{Masterarbeit}
\newcommand{\kind}{Masters thesis}

\newcommand{\kitg}{Karlsruher Institut für Technologie}
\newcommand{\kit}{Karlsruhe Institute of Technology}
\newcommand{\kitlongg}{KIT -- Universität des Landes Baden-Württemberg und nationales Forschungszentrum der Helmholtz-Gesellschaft}
\newcommand{\kitlong}{KIT -- University of the State of Baden-Wuerttemberg and National Laboratory of the Helmholtz Association}
\newcommand{\facultylongg}{An der Fakultät für Informatik}
\newcommand{\facultylong}{At the faculty of Computer Science}
\newcommand{\instituteg}{Institut für Anthropomatik und Robotik}
\newcommand{\institute}{Institute for Anthropomatics and Robotics}
\newcommand{\groupg}{Computer Vision for Human-Computer Interaction Research Group}
\newcommand{\group}{Computer Vision for Human-Computer Interaction Research Group}
\newcommand{\stateheadg}{Erklärung}
\newcommand{\statehead}{Statement of Authorship}
\newcommand{\statementg}{Hiermit erkläre ich, die vorliegende Arbeit selbständig erstellt und keine anderen als die 
angegebenen Quellen verwendet zu haben.}
\newcommand{\statement}{I hereby declare that this thesis is my own original work which I created without illegitimate help 
by others, that I have not used any other sources or resources than the ones indicated and that due acknowledgement is 
given where reference is made to the work of others.}
\newcommand{\thesisofg}{\kindg\\von}
\newcommand{\thesisof}{\kind~of}
\newcommand{\durationlabg}{Bearbeitungszeit}
\newcommand{\durationlab}{Duration}
\newcommand{\revieweronelabg}{Erstgutachter}
\newcommand{\revieweronelab}{Reviewer}
\newcommand{\reviewertwolabg}{Zweitgutachter}
\newcommand{\reviewertwolab}{Second reviewer}
\newcommand{\advisorlabg}{Betreuender Mitarbeiter}
\newcommand{\advisorlab}{Advisors}

\makeatletter
\DeclareRobustCommand\onedot{\futurelet\@let@token\@onedot}
\def\@onedot{\ifx\@let@token.\else.\null\fi\xspace}
\def\vs{vs\onedot}
\def\eg{\emph{e.g}\onedot} \def\Eg{\emph{E.g}\onedot}
\def\ie{\emph{i.e}\onedot} 
 
 \def\vs{\emph{vs}\onedot}
 
\def\etal{\emph{et al}\onedot}
\makeatother

\def\front{\emph{front}\xspace} 
\def\side{\emph{side}\xspace}
\def\back{\emph{back}\xspace}
\def\left{\emph{left}\xspace}
\def\right{\emph{right}\xspace}
\def\viewunit{\emph{view unit}\xspace}
\def\viewunits{\emph{view units}\xspace}
\def\logits{\emph{logits}\xspace}
\def\viewpredictor{\emph{view predictor}\xspace}

\newcolumntype{L}[1]{>{\raggedright\let\newline\\\arraybackslash\hspace{0pt}}m{#1}}
\newcolumntype{C}[1]{>{\centering\let\newline\\\arraybackslash\hspace{0pt}}m{#1}}
\newcolumntype{R}[1]{>{\raggedleft\let\newline\\\arraybackslash\hspace{0pt}}m{#1}}

\newcommand{\f}[1]{\textbf{#1}} 
\newcommand{\np}[0]{$^\dag$}    

\newcommand{\mcl}[1]{\multicolumn{2}{|l|}{#1}}

\begin{document}

\bstctlcite{IEEEexample:BSTcontrol}

\selectlanguage{english}

\frontmatter
\pagenumbering{roman}
\include{titlepage}
\include{title}

\blankpage
\include{content/abstract}
\include{content/acknowledgement}

\tableofcontents
\blankpage

\mainmatter
\pagenumbering{arabic}
\normalem
\include{content/introduction}

\include{content/related}

\include{content/methodology}

\include{content/evaluation}

\include{content/conclusion}

\cleardoublepage
\phantomsection
\addcontentsline{toc}{chapter}{\bibname}

\iflanguage{english}
 {\bibliographystyle{IEEEtranSA}}	
 {\bibliographystyle{babalpha-fl}}	


\bibliography{thesis}

\end{document}

%% file: titlepage.tex

\newcommand{\diameter}{20}
\newcommand{\xone}{-15}
\newcommand{\xtwo}{160}
\newcommand{\yone}{25}
\newcommand{\ytwo}{-240}

\begin{titlepage}
\begin{tikzpicture}[overlay]
\draw[color=gray]  
 		 (\xone mm, \yone mm)
  -- (\xtwo mm, \yone mm)
 arc (90:0:\diameter pt)
  -- (\xtwo mm + \diameter pt , \ytwo mm)
	-- (\xone mm + \diameter pt , \ytwo mm)
 arc (270:180:\diameter pt)
	-- (\xone mm, \yone mm);
\end{tikzpicture}
\begin{textblock}{10}[0,0](4,2.5)
	\iflanguage{english}{\includegraphics[width=.3\textwidth]{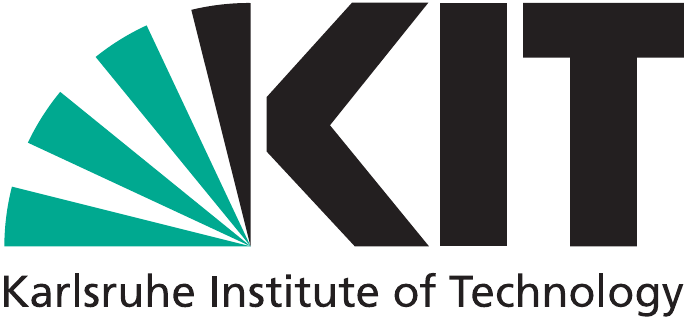}}{\includegraphics[width=.3\textwidth]{logos/kitlogo_de.pdf}}
\end{textblock}
\changefont{phv}{m}{n}	
\vspace*{3.3cm}
\begin{center}
	\Huge{\mytitle}
	\vspace*{2cm}\\
	\Large{
		\iflanguage{english}{\thesisof}{\thesisofg}
	}\\
	\vspace*{1cm}
	\huge{\dtauthor}\\
	\vspace*{1cm}
	\Large{
		\iflanguage{english}{\facultylong}{\facultylongg}
		\\
		\iflanguage{english}{\institute}{\instituteg}
	}
	\end{center}
	\vspace*{1cm}
	\vspace*{1cm}
\Large{
\begin{center}
\begin{tabular}[ht]{l c l}
  \iflanguage{english}{\revieweronelab}{\revieweronelabg}: & \hfill  & \reviewerone\\
  \iflanguage{english}{\reviewertwolab}{\reviewertwolabg}: & \hfill  & \reviewertwo\\
  \iflanguage{english}{\advisorlab}{\advisorlabg}: & \hfill  & \advisor\\
\end{tabular}
\end{center}
}

\vspace{2cm}
\begin{center}
\large{\iflanguage{english}{\durationlab}{\durationlabg}: \timestart \hspace*{0.25cm} -- \hspace*{0.25cm} \timeend}
\end{center}

\begin{textblock}{10}[0,0](4,16.65)
\tiny{ 
	\iflanguage{english}{\kitlong}{\kitlongg}
}
\end{textblock}

\begin{textblock}{10}[0,0](14,16.6)
\large{
	\textbf{www.kit.edu} 
}
\end{textblock}

\end{titlepage}

%% file: title.tex
\phantomsection
\pdfbookmark[0]{Title}{sec:title}
\dttitle{\dttit}{\dtauthor}{\dtaddress}

\dtlegal{\dtauthor}{\dtdateother}

%% file: content/abstract.tex

\thispagestyle{empty}
\chapter*{Abstract}

Person re-identification (re-id) is the task of recognizing and matching persons at different locations recorded by cameras with non-overlapping views.
One of the main challenges of re-id is the large variance in person poses and camera angles since neither of them can be influenced by the re-id system.
In this work, an effective approach to integrate coarse camera view information as well as fine-grained pose information into a convolutional neural network (CNN) model for learning discriminative re-id embeddings is introduced.
In most recent work pose information is either explicitly modeled within the re-id system or explicitly used for pre-processing, for example by pose-normalizing person images.
In contrast, the proposed approach shows that a direct use of camera view as well as the detected body joint locations into a standard CNN can be used to significantly improve the robustness of learned re-id embeddings.
On four challenging surveillance and video re-id datasets significant improvements over the current state of the art have been achieved.
Furthermore, a novel reordering of the MARS dataset, called X-MARS is introduced to allow cross-validation of models trained for single-image re-id on tracklet data.

%% file: content/acknowledgement.tex

\thispagestyle{empty}
\chapter*{Acknowledgement}

My appreciation goes to Prof. Dr.-Ing. Rainer Stiefelhagen and my advisors M. Saquib Sarfraz and Arne Schumann for providing me the opportunity to work on the interesting subject of person re-identification with convolutional neural networks.
This subject is not only of great interest for society but also a tremendous example of the potential of neural networks and what can be achieved in this field of research. 

Many thanks goes to my colleagues at arconsis IT-Solutions GmbH for all the motivating and inspiring conversations and the freedom to do this work while also working on great projects with them.
Moreover, I want to thank Alexander Frank and Wolfgang Frank for introducing me to the world of neural networks through a project in which they have been a great team lead and boss, respectively.

Special thanks go to Katja Leppert and Valentin Zickner for giving valuable input and proofreading this work to improve it further. 

I would also like to give thanks to my family and friends for all their support. Especially, I want to thank my older brother Christian Eberle, who was the one introducing me to the world of computer programming.

%% file: content/introduction.tex

\chapter{Introduction}
\label{ch:Introduction}

\begin{figure}
\centering
\includegraphics[width=\columnwidth]{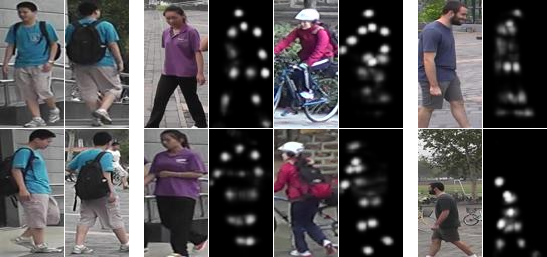}
\caption{Example images from Market-1501, DukeMTMC-reID and PRW datasets showing that camera perspective and body pose can vary significantly between different person images. While a different view angle might show different aspects of a person (\eg a backpack), a different body pose can change the location of local features (\eg the location of an arm or a leg). Furthermore, pose information can help to guide the attention of a person re-id system towards relevant image regions in cases of mis-alignments.}
\label{fig:motivation}
\end{figure}

Person re-identification (re-id) in non-overlapping camera views has attracted more and more attention during recent years as displayed by the large number of works released in this area. 
With the increase in available computational power and large datasets fueling the rise of Convolutional Neural Networks (CNNs), large improvements have been seen within the field computer vision in general and this challenging matching problem in particular.

Applications for this technology range from automated surveillance of locations like malls and airports or of large events with increased security requirements to
person tracking \eg in restricted environments like embassies or laboratories with strong security restrictions.

Most of the challenges for person re-id arise from the uncontrolled environment.
Cameras cannot necessarily be positioned with an overlapping field-of-view, have varying distances towards recorded persons, and are produced by different manufacturers with different specifications and characteristics.
For example, the resolution of the camera and thus the resolution of a detected person as well as the focal length, color balance, and other camera sensor characteristics can be different.
Moreover, since cameras are positioned in different places (\eg indoors \vs outdoors) and images are recorded at different times during the day, lighting and other external influences can vary significantly.
Furthermore, since storage of uncompressed video data of many cameras quickly generates very large amounts of data, compression techniques are used, causing artifacts in the images.

On top of these variations, the pose of a recorded person is often mostly unconstrained.
This not only includes the view angle a person has towards the camera (\eg a person can walk towards, away or orthogonal to the camera) but also the full body pose (\eg for a walking person legs and arms are moving).
Due to the large impact of pose variations on the visual appearance of a person, these two aspects probably have the largest impact on re-id.

Most previous solutions utilizing CNNs try learning a person's global appearance by either applying a straightforward classification loss function or by using a metric learning loss function. 
In the case of the classification loss, the loss is applied to a classification layer during training, while internal embeddings are used during evaluation and compared with a distance function.
Hence, training is not directly focused on the actual objective (\ie finding discriminative embeddings) but such embeddings are found rather implicitly. 
In contrast, with metric learning the embeddings and therefore the actual objective are optimized directly.

In order to improve learning of local statistics, these approaches have also been applied to local image regions like horizontal stripes and grids \cite{li2014deepreid,cheng2016person}. 
Due to non-overlapping camera views, varying view angles and varying person poses, re-id faces the challenge that there is no implicit correlation between local regions of the image (see Figure \ref{fig:motivation}).
However, this can be established by exploiting explicit full body pose information for alignment or for matching detected body parts locally \cite{zhao2017spindle, zhao2017deeply}. Utilizing the full body pose as additional information for local or global person description has been shown to strongly benefit person re-id.

In this thesis, two practical ways to extend common CNN architectures allowing to incorporate coarse pose (\ie the view angle a person has towards the recording camera) and fine-grained pose information (\ie joint locations) are presented.
It is shown that providing a standard CNN architecture with the person's joint locations as additional input channels helps to increase person re-id accuracy.
Likewise, learning and combining view-direction-specific feature maps improves the embeddings as well.
Additionally, it is demonstrated that combining both of these extensions to embed coarse and fine pose information into a standard CNN architecture improves the re-id embeddings further.
In all, exploiting view and pose information greatly benefits re-id performance while only using a simple classification loss. 

With person re-id systems becoming more and more accurate, real-world considerations gain more importance.
One such aspect is the use of tracklets instead of single images, because most person re-id systems are based on video sequences where often several images can be extracted for a single person.
Because of the increased amount of information and variance contained in the time-series tracklets, person re-id performance can be improved.
One large dataset allowing the evaluation of re-id on tracklets is MARS \cite{MarsDataset}.
However, as annotation of tracklet data is especially time-consuming and thus expensive, the usage of a single-image re-id system for detection of tracklets has large potential for cost and training time reduction.
Unfortunately, MARS cannot be used for cross-evaluations with the related single-image Market-1501 dataset \cite{Market1501Dataset}, because the training and test sets of both datasets overlap.
To alleviate this problem the X-MARS dataset is proposed, which is a reordering of the MARS dataset to remove the overlap between Market-1501's and MARS' training and test sets.

In summary the contributions of this thesis are fourfold.
\begin{itemize}
\item[1] Two new CNN embeddings are proposed incorporating coarse view and fine-grained pose information. 
Furthermore, both of these can be combined to form a combined embedding and it is shown that the two pose information complement each other.
\item[2] On three challenging person re-id datasets, the pose-sensitive person re-id model sets a new state of the art.
\item[3] To enable cross evaluation of networks trained on a single-image dataset on tracklet datasets, the X-MARS reordering is introduced, allowing further evaluation of real-world considerations.
\item[4] The proposed embeddings are further evaluated under several settings relevant to real world applications, including image-to-video re-id on the proposed X-MARS benchmark, scalability with very large gallery sizes, and robustness to errors resulting from an automated person detection.
\end{itemize}

%% file: content/related.tex

\chapter{Related Work}
\label{ch:RelatedWork}

The challenging task of person re-identification has a long history of approaches developed to tackle it.
Over time, applied methods and principles have shifted from the use of handcrafted features, as discussed in Section~\ref{section:generalReId}, towards an automated learning with convolutional neural networks, which are detailed in Section~\ref{section:cnnReId}.
Since re-id has to handle drastically varying pose conditions, the detection and usage of the presented pose to improve re-id has been subject to previous research as well.
In Section~\ref{section:poseReId} focus is put on these works while in Section~\ref{section:related:poseEstimation} approaches of getting pose estimations are discussed.

On top of these works on improving the way person re-id is done, re-ranking methods have gained large interest during recent years.
Since these re-ranking methods allow to improve the re-id results further, two recent unsupervised approaches are presented in Section~\ref{section:reRanking}.

\section{General Re-Identification}
\label{section:generalReId}

While typically re-identification of persons is done by using classical biometric characteristics like a person's face or fingerprint, this is impractical for video surveillance scenarios due to low resolutions of the recorded video data and the unconstrained environments.
Instead, various aspects like clothing and the overall appearance are utilized \cite{doretto2011appearance} to create illumination and pose impervious representations.

The term ``person re-identification'' was probably first used in the work of Zajdel, Zivkovic and Kröse \cite{zajdel2005keeping}.
In their paper they try to re-identify a person when it leaves the field of view of one camera and enters the field of view of another.
To achieve this, they assume a unique hidden label for all persons and create a dynamic Bayesian network encoding the probabilistic relationships between the labels and features (\eg spatial and temporal cues as well as color) gathered from the tracklets.

In the work of Cheng \etal \cite{cheng2006matching}, effects of variable illumination and camera differences are tackled with a cumulative color histogram transformation on the segmented object.
An incremental major color spectrum histogram is then used to form a representation of the object that is able to handle small appearance changes.

Wang \etal \cite{wang2007shape} propose a framework including an appearance model to handle similarities between deformable parts of an object.
With this approach they are able to cover the spatial distribution of the object parts' appearance relative to each other.

To improve re-id with varying viewpoints, Gray and Tao \cite{gray2008viewpoint} propose and ensemble of localized features (ELF). 
Furthermore, instead of handcrafting the complete features to represent a person for re-id matching, they hand-design a feature space and utilize machine learning to find a class specific representation to build a discriminative recognition model. 
By doing so, they are able to combine many simple handcrafted features for the final representation.

To analyze features of a person and prevent the influence of the background, many approaches require the pedestrian to be separated from the background. 
Bouwans \etal \cite{bouwmans2008background} and Dollar \etal \cite{dollar2012pedestrian} give comprehensive overviews of methods used to detect pedestrians and separate fore- and background.

To better handle changes in appearance caused by changed poses of recorded persons, a large amount of part-based body models have been developed (see \cite{satta2013appearance} for a comprehensive listing).
While body models allow the combination and extraction of global and local appearance features, spatial relation and feature combination is often difficult.
Furthermore, large changes in pose and orientation, which are typical for the re-id task with non-overlapping camera views, often pose challenges.

Besides appearance focused re-identification approaches other cues have been investigated as well.
While these often suffer some intrinsic limitations, they can still be used as additional hints to improve re-id matching.
When video data (\ie multiple sequential images of a person) is available, recurrent patterns of motion can be analyzed \cite{stevenage1999visual}.

A different way is chosen by Layne \etal \cite{layne2012person} where persons are re-identified based on semantic attributes like \eg if they are wearing sandals, a backpack or shorts.
Their system focuses on learning mid-level semantic features and can be used to complement other approaches.

A work of Han \etal \cite{man2006individual} proposes a technique called Gait Energy Image to characterize a person's walking behavior.
In particular, they normalize, align, and average sequences of foreground-only silhouettes from a single walking period.
Additionally a Principal Component Analysis is used to reduce the dimensionality of the extracted features.
While requiring processing of video data, motion analysis can overcome the limitations posed by only observing appearance for re-id matching.

Another additional cue can be provided by anthropometry, the measurement of physical body features \cite{roebuck1975engineering}.
These techniques try to estimate \eg a person's height, leg length or eye-to-eye distance. 
However, for these measurements, body landmarks have to be localized, often implying costly calculations or special hardware not available in the general re-id scenario.

Most of these approaches have in common that they generate features which need to be compared to do the actual matching for re-identifying a person.
Metric learning methods try to optimize the metrics used for these comparisons.
An extensive study of such methods was done by Yang and Jin in \cite{yang2006distance}.

Metric learning techniques are mainly categorized into supervised learning and unsupervised learning as well as local learning and global learning approaches.
For example, the Mahalanobis distance follows the idea of global metric learning to keep all vectors of the same class close and push vectors of other classes further away.
Xing \etal \cite{xing2003distance} formulate a convex programming problem to optimize the distance calculation.
The popular and more recent KISSME \cite{koestinger2012large} method formulates the similarity of two features as a likelihood ratio test.
Additional principle component analysis (PCA) is applied to remove redundant dimensionality.

\section{Re-Identification with Convolutional Neural Networks}
\label{section:cnnReId}

During recent years many state-of-the-art results in the field of person re-identification have been achieved by using Convolutional Neural Networks (CNNs) to learn feature embeddings automatically.
In contrast to methods presented in the previous section in which features are handcrafted, CNNs implement a data-driven approach to find the best feature extractors with machine learning based on the data presented during training.
This enables the use of a much larger number of features as they are found automatically instead of being developed manually.
The use of CNNs for image processing was sparked by the huge performance gains shown by Krizhevsky \etal \cite{krizhevsky2012imagenet} with their AlexNet in the ImageNet competition.

CNNs themselves are a special type of artificial neural networks gaining their strength by learning filters used to extract features from their input.
Feature extraction is done similarly to the way of classical approaches by convolving the filters with the input, which results in applying the filter for every location in the input.

Although limiting the learnable weights to filters of a much smaller size than the input limits, the neural network's degree of freedom (\eg in comparison to a fully connected layer) actually allows a more bounded and thus stable way of learning the feature extraction.
While motivation for this restriction comes from image processing, where it is reasonable to assume that a feature (\eg an edge) can be calculated the same way all over the image, they have also been applied to time-series and speech recognition \cite{lecun1995convolutional}.
Furthermore, by stacking convolutional layers, one can create a hierarchy of learnable feature extractors capable to extract more and more complex features. 
While each level of features is only created from a local neighborhood, deeper layers take larger and larger fields of view into account. 
Moreover, by fusing low-level features, higher-level features can be created with every layer.
All in all, CNNs apply the same ideas of extracting local features from images to combine them into meaningful high-level features like it has been done in manually designed systems for a long time. 
Their advantage however is that the best filters for the given task can be automatically found during training with back-propagation instead of being handcrafted.

When training neural networks a loss function is used which can be optimized by the back-propagation algorithm. 
For person re-id, there are two main groups of how to optimize the embeddings used for matching via a distance function. 
On the one side, one can add a classification layer on top of the embeddings and train this classification layer \eg with a softmax-cross-entropy loss.
By doing so, the embeddings are not trained directly and thus the objective of achieving a good re-id is not trained directly.
In contrast, when using a triplet loss \cite{cheng2016person,trinet}, one can directly optimize the embeddings. 
Here the objective is to minimize the distance between samples of the same person and increase the distance between different persons.
However, triplet loss requires to mine good triplets for the learning process to work best.

Deep learning models have first been used for person re-id by Yi \etal in \cite{yi2014deep} and by Li \etal in \cite{li2014deepreid}.
Yi \etal \cite{yi2014deep} partition the input image into three overlapping horizontal stripes which are processed by two convolutional layers before being fused by a fully connecting layer at the end. 
Feature vector comparison is done with the cosine distance.
In contrast, the siamese architecture proposed by Li \etal \cite{li2014deepreid} processes two images directly to compare them.
For this, a patch matching layer is used to fuse horizontal stripes of the two different images to fuse their features which are then fused with fully connected layers towards a same person / different person classification.

Ahmed \etal \cite{ahmed2015improved} improve the siamese model by comparing features of neighboring locations between the input images.
Wu \etal \cite{wu2016personnet} use smaller sized convolutional filters allowing them to deepen the network.
Long short-term memory (LSTM) layers are used in \cite{varior2016siamese} to process parts of the person image sequentially allowing the LSTM to learn spatial relations between these parts.
Cheng \etal \cite{cheng2016person} extend the idea of siamese networks to networks processing three images at once and introduce triplet loss for direct metric learning of the embeddings.
Hermans \etal continue this idea providing a simpler way for training with \cite{trinet}.

\section{Person Re-Id Using Pose Information}
\label{section:poseReId}
This section focuses on approaches utilizing a degree of pose information to improve person re-id.

The popular SDALF approach by Farenza \etal \cite{farenzena2010person} is based on multiple phases.
At first it separates the person in the image from the background to then search for asymmetry and symmetry axes in the pedestrian's image.
These symmetry axes are used to split the image and extract high entropy segments of the image from all tiles correlating to different body parts.
Afterwards, features are extracted from the equalized foreground image and used for matching.

Cho \etal \cite{cho2016improving} propose the usage of views for multi-shot matching problems where image sequences are compared.
They define four views (front, back, right and left) and estimate the views of gallery and query images to weight them to emphasize same-view person images.
This idea is based on the observation that when a front query image is compared with a back gallery image the correlation is usually less reliable as when comparing a front query to a front gallery image.
Accordingly, comparisons of non-matching views are weighted less than comparisons of matching views.

An approach focused on more fine-grained pose information was first introduced by Cheng \etal \cite{cheng2011custom, cheng2014person}. 
They adapt Pictorial Structures to find and extract body parts to match their descriptors.
By doing so, explicit focus is put on the body parts.
For multi-shot matching, they propose a Custom Pictorial Structure to better learn the appearance of an individual by exploiting the information provided by multiple shots leading to an improvement in body part detection and thus person re-id.

The significant successes of CNN architectures in the context of re-id have lead to multiple works directly including pose information into a CNN-based matching. 
Zheng \etal \cite{poseInvariantEmbedding} use a of-the-shelf CNN based pose estimator to locate body joints. 
Based on the body joints, body parts are cut out and put together to a standardized PoseBox image.
They then feed two CNN branches of their network with the original person image and the normalized PoseBox image and combine the resulting CNN feature maps with the confidence of the pose estimation to form a single deep re-id embedding.

A similar approach is developed by Su \etal \cite{su2017pose}.
A sub-network first estimates fourteen pose maps which are then used to localize body joints as well as to crop and normalize the body parts based on the joints locations.
Again, original and normalized images are fed into a CNN network with two branches to learn local as well as global person representations which are fused for a final embedding.
In contrast to Zheng \etal, this approach can be learned end-to-end as it integrates the pose estimation and alignment into the network instead of relying on an external pose estimator.

The Spindle Net CNN proposed by Zhao \etal \cite{zhao2017spindle} uses pose information for a multi-staged feature decomposition for seven body parts.
Features for the body parts are then fused in a tree-structured competitive network structure to enable the CNN to incorporates macro- and micro-body features.

A different concept is proposed by Rahimpour \etal \cite{rahimpour2017person}.
Here, visual attention maps generated via a sub-network are used to guide the actual deep CNN used for re-id matching. 
Furthermore, a triplet loss is applied for training the embedding's objective directly.

Zhao \etal \cite{zhao2017deeply} create a deep CNN learning person part-aligned representations.
Their model decomposes the person image into body regions and aggregates the calculated similarities between corresponding regions as the overall matching score.
For training, they also employ a triplet loss. 

In contrast to the approach presented in this work, all of these works mostly rely on fine-grained body pose information. 
Moreover, these methods either utilize pose information by explicitly transforming or normalizing the input images or by explicitly modeling part localization in their architecture. 
Contrarily, this work's approach relies on confidence maps of body joint locations generated by a pose estimator, which are simply added as additional input channels alongside the input image. 
This allows the network a maximum degree of flexibility and leaves it to the network to learn how to use the confidence maps best and which body parts are most reliable for re-id.
On top of this fine-grained pose information, more coarse pose information is exploited as well. 
It turns out that this coarse pose is even more important for re-id and can efficiently be used to improve a system's performance.

\section{Pose Estimation}
\label{section:related:poseEstimation}

Several previously described methods as well as this work require to retrieve pose information of a detected person solely from the image.
This section describes some of these methods including the DeeperCut \cite{DeeperCut} model utilized in this work.

Early work on pose estimation was done by Jiang \etal \cite{jiang2008global}. 
Their method is based on integer linear programming to compose potential body part candidates into a valid configuration.

Eichner \etal \cite{eichner2010we} propose a multi-person pose estimation based on pictorial structures explicitly modeling interactions and occlusions between people.
Due to a combined processing of all people in the image, they can improve pose estimation especially in the case where multiple people are standing close to each other.
However, only upper body part poses are estimated.

A combined pose estimation and segmentation method is described by Ladicky \etal \cite{ladicky2013human} using a greedy approach to add single person hypotheses to the joint objective step by step.

The DeepCut model presented by Pishchulin \etal \cite{deepCut} follows a joint approach to detect persons and estimated their body pose together.
The result candidates proposed by their CNN-based part detector are grouped into valid configurations with integer linear programming respecting appearance and geometrical constraints.

With DeeperCut Insafutdinov \etal \cite{DeeperCut} improve the ideas of the DeepCut model with improved body part detectors, image-conditioned pairwise terms and an incremental optimization strategy. 
With these improvements they gain better detection performance while improving the detection speed.

\section{Re-Ranking Methods}
\label{section:reRanking}

In recent years, more and more attention has been drawn by re-ranking techniques in the field of person re-id. 
With re-id being a retrieval process, re-ranking can significantly improve accuracy.

The work of Zhong \etal \cite{zhong2017re} provided a strong impulse to the use of re-ranking for person re-id.
Their k-reciprocal encoding is based on the idea that if a gallery image is similar to the query image within the k-reciprocal neighbors, it is more likely to be a true match.
To allow efficient computing of the k-reciprocal neighbor distances they calculate the Jaccard distance with Sparse Contextual Activation (SCA).
In the end, the original distance and the calculated Jaccard distance are combined to retrieve the final ranking.

A different approach is taken by Sarfraz \etal in \cite{pse-ecn} for their Expanded Cross Neighborhood (ECN) re-ranking.
ECN works by summing the distances of immediate neighbors of each image with the other image's results achieving the current state of the art.
Moreover, in contrast to k-reciprocal re-ranking they do not strictly require rank list comparisons and can work with a simple list comparison measure.

In this thesis, results are also provided re-ranked with both, k-reciprocal and ECN re-ranking setting a new state of the art for re-ranked person re-id results.

%% file: content/methodology.tex

\chapter{Pose-Driven Deep Models}
\label{ch:pse}

In the setting of person re-id a person's appearance is greatly affected by their view angle in relation to the camera recording the image.
This is especially important in the case of non-overlapping camera views where the view angles can be inherently different between cameras.
Furthermore, the visual appearance and body part locations in the image are significantly influenced by a person's pose. For example, the positioning of legs and arms can largely differ over time when a person is walking. 
In order to enable the model to handle both of these challenges an explicit modeling of a person's pose can be helpful.

In this section two pose-sensitive extensions are introduced and explicitly incorporated into two existing neural network architectures. 
In Section~\ref{section:pse:views} the view angle of a person in regard to the camera is utilized, whereas in Section~\ref{section:pse:pose} the body pose information is exploited to guide a network's attention and enable it to focus on parts of the person's body.
Moreover, in Section~\ref{section:pse:pse} a combination of both extensions is proposed to further improve detection performance.

\section{View Information}
\label{section:pse:views}

\begin{figure}
\centering
\includegraphics[width=\columnwidth]{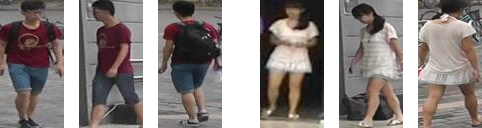}
\caption{Three different views (\front, \side, \back from left to right) of two persons towards the camera. The images show two identities of the Market-1501~\cite{Market1501Dataset} training set.}
\label{fig:views_on_market}
\end{figure}

A person's angle towards the camera has a great effect on their appearance, as shown in Figure~\ref{fig:views_on_market}.
For example, the body shape changes significantly between \front and \back \vs \side views. 
Likewise, while \front views mostly contain visible faces, \side views often only contain portions of the face and \back views contain almost no faces at all. 
The same applies for clothing or items persons are carrying (\eg backpacks or handbags). 

Thus, it seems reasonable that re-id accuracy can improve when images of different view angles are handled differently.

If the view angle of each image were known, a naive approach would be to train a CNN with multiple full-depth branches, one for each discretization of a person's view angle towards the camera and combine the results in the end.
Since low- and mid-level features required to detect top-level features will be similar between these branches, one can combine them into a single network where only a part at the end (in the following referred to as \viewunit) is replicated and activated depending on the image's view angle.
However, because the test time images will not have labels for their view angle, the network needs to detect the angle in order to activate the correct \viewunit.

Inspired by a recent work of Sarfraz \etal \cite{SarfrazPedestrian17} on semantic attribute recognition, a ternary \viewpredictor side-branch is included into the base person re-id CNN.
View probabilities are calculated by applying the softmax function to the result of the \viewpredictor side-branch.
Furthermore, the tail part of the CNN is replicated multiple times and the resulting feature maps of each \viewunit are weighted with the view probabilities of the aforementioned side-branch.
The weighted feature maps are then summed and fed into the final layers of the original architecture to create the actual embeddings used for person re-id.
By weighting the view units' feature maps with the view prediction, the gradient flowing through them is modulated. 
If, for example, a frontal view is detected by the view predictor branch, the corresponding view unit will most strongly contribute to the final embedding and thus mainly this \viewunit will be adjusted during back-propagation to better describe front facing images while the gradient flow is blocked or reduced for the other \viewunits.
To achieve a more robust representation, the weighting is applied to full feature maps, which are then fused and fed into the final layers.
In contrast to the feature fusion used in this work, Sarfraz \etal originally apply a decision-level fusion to generate their final results.

\begin{figure}[b]
\centering
\includegraphics[width=\columnwidth]{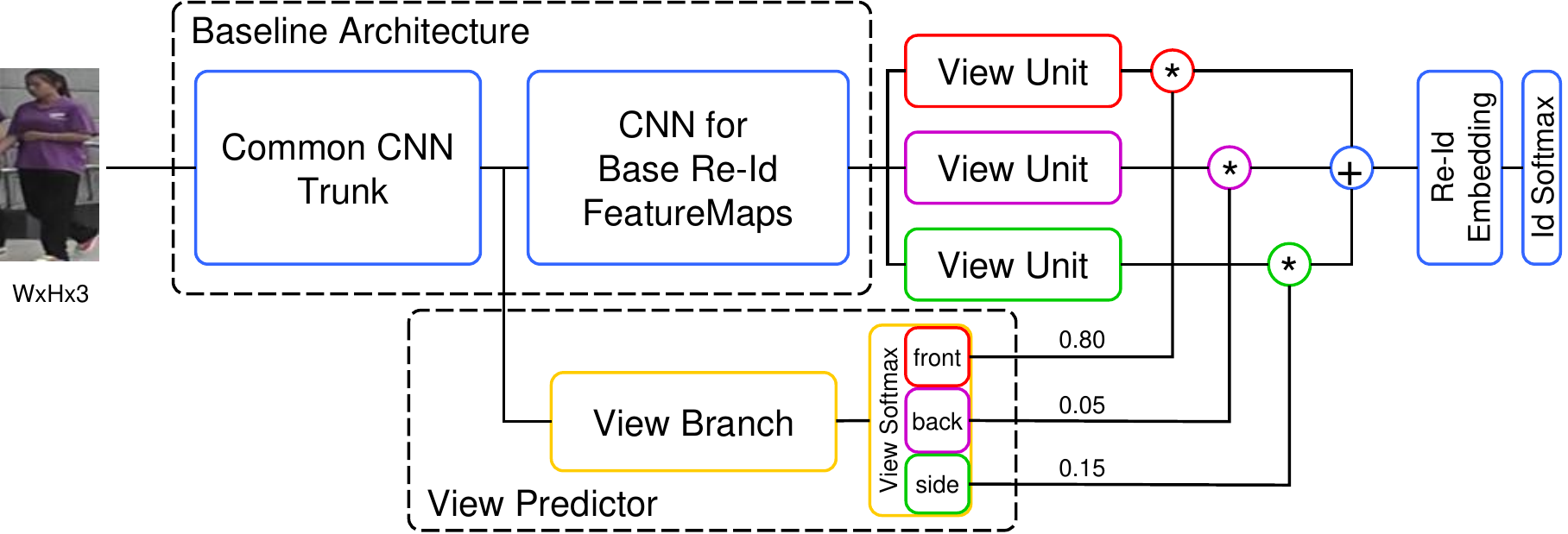}
\caption{View model network extension. The \viewpredictor is branched off from within the base model and used to predict view probabilities which are then used to weight the \viewunits' feature maps.}
\label{fig:viewModel}
\end{figure}

Figure \ref{fig:viewModel} illustrates the resulting view network extension. 
The used baseline CNN architecture is split into three parts.
The first layers are a common CNN trunk preceding all remaining components of the view model.
Afterwards, the remaining part of the baseline model, except for its last block, is added to form the base re-id feature maps. 
These base feature maps are then used by three replicas of the last block.
Note that these \viewunits are independent units not sharing weights with each other and are thus able to learn different features.
To be able to predict the person's view, the \viewpredictor is added after the common CNN trunk forming a side branch to the main network.
The softmax activation of the \viewpredictor is then used to weight the feature maps of the \viewunits before summing and feeding them into the re-id embedding layer.
Finally, a softmax classification layer is added to train the model.
A detailed illustration of the \viewpredictor, the \viewunits and their connection is given for the ResNet-50 base architecture \cite{ResNet} in Figure~\ref{fig:pseResNetModel} on page~\pageref{fig:pseResNetModel}.

\section{Full Body Pose Information}
\label{section:pse:pose}
\begin{figure}
\centering
\includegraphics[width=\columnwidth]{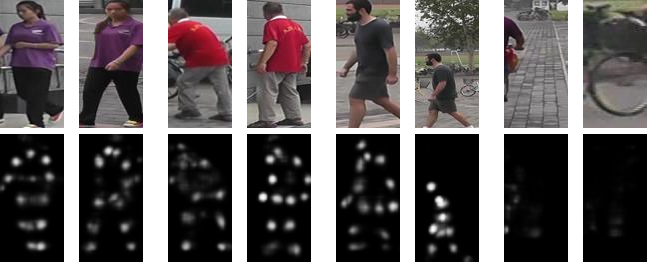}
\caption{The upper row shows three identities of the Market-1501 training set and two false detections. While the images of each person are recorded at roughly the same view angle, their appearance varies due to their pose and image alignment. The lower row shows the joint locations as detected by the DeeperCut pose model. For the false positive detection images, the pose maps show almost no activation.}
\label{fig:poses_on_market}
\end{figure}

Like the view angle of a person towards the recording camera significantly changes the appearance, a person's pose and their alignment in the image can have an impact as well. 
Figure~\ref{fig:poses_on_market} shows how the location and correlation between body parts change with different body poses and bad person detection alignments. 
Hence, providing the network with the full body pose of the person acts to guide the network's attention towards the different body parts, regardless of their positioning.

To acquire the fine-grained pose information of a person, the off-the-shelf DeeperCut \cite{DeeperCut} pose estimator is used, which estimates the location of 14 main joint keypoints.
Figure \ref{fig:poses_on_market} shows the final feature layers of the DeeperCut network's detections.
For better visualization, the 14 channels have been combined into a single-channel gray image by applying a \emph{maximum} operation across all channels.

Usually, the DeeperCut model's final feature layers are used to find the coordinates with the highest value via an \emph{argmax} operation. 
In contrast, the proposed pose CNN is directly provided with DepperCut's final feature layers. 
One reason for this is the fact that the images might show incomplete persons, missing some body parts.
By providing the feature maps instead of coordinates, a body part detection (\ie a hard decision) is not enforced where there might not even be a body part in the image.
Additionally, this helps to compensate difficult pose detection cases where the DeeperCut model's detection has high uncertainty.
In these cases the confidence maps do not show a clear hot spot but \eg a flat confidence over a wide area, as visualized by the two false detections on the right in Figure \ref{fig:poses_on_market}.
Additionally, the 14 confidence maps provide a much more detailed view of the body pose than simple coordinates would. 
Thus, by feeding the confidence maps as additional channels alongside the three color channels of the input image, a way to guide the network's attention is provided while leaving it to the network to learn how best to utilize the full body pose information.

In contrast, \cite{poseInvariantEmbedding} use pose estimation to generate an artifical image with explicitly aligned body parts, called \emph{PoseBox}.
This \emph{PoseBox} is then fed into the network alongside the original image and the confidence of the pose estimation.
In Section \ref{section:evaluation:PSEvsPIE} the differences between this explicit modeling and the way proposed by this work are evaluated.

While in this thesis the DeeperCut network is employed, it is important to note that any other pose estimator could be used as well. 
Furthermore, the number and type of keypoints could also be changed.

\begin{figure}
\centering
\includegraphics[width=\columnwidth]{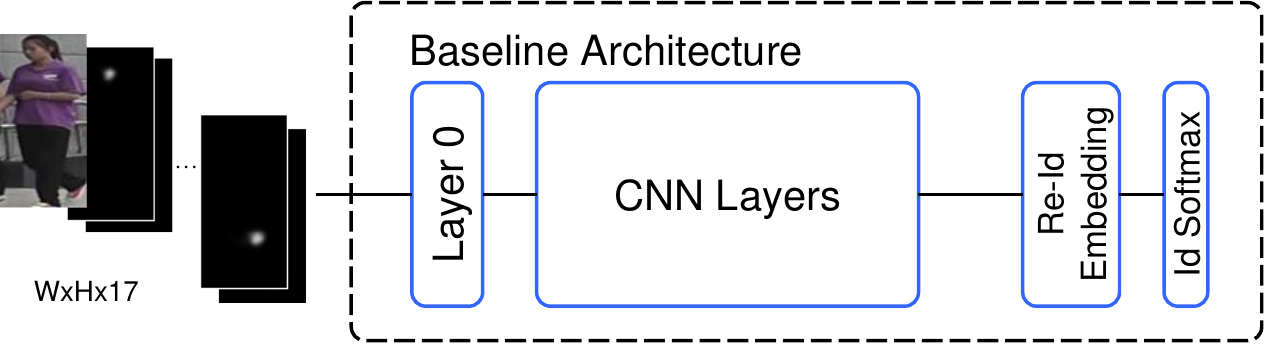}
\caption{Pose model network extension. The 14 pose maps generated by the DeeperCut pose estimator are fed into the network alongside the input image's three color channels.}
\label{fig:poseModel}
\end{figure}

Figure \ref{fig:poseModel} visualizes the main difference between the baseline model and the pose model. 
All of the base network architecture's layers remain unchanged, except for the first one.
Since the network is fed with the original three color channels of the input image and the 14 joint location maps generated by the DeeperCut pose estimator, the first layer has to be changed to accept 17 input channels instead of three.

\section{Pose-Sensitive Embedding (PSE)}
\label{section:pse:pse}

\begin{figure}
\centering
\includegraphics[width=\columnwidth]{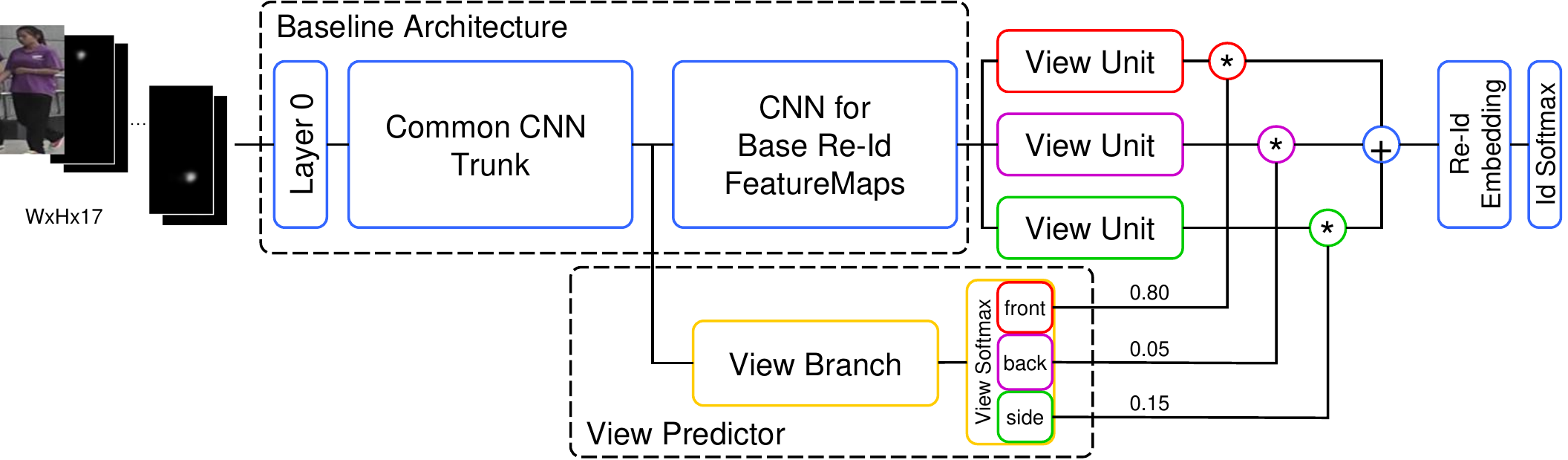}
\caption{PSE model combining the \viewpredictor with the usage of pose maps.}
\label{fig:pseModel}
\end{figure}

While both of the previous extensions incorporate view information into the network to improve its accuracy, combining them could provide additional benefit. 
On the one hand, this is because the coarse view angle information and the fine-grained joint locations are very different types of information. 
Whereas the view angle causes a change in the overall appearance, a changed body pose (\eg a lifted arm or shift of the legs when walking) is a more localized change and thus influences rather a part of the image than the whole image.
On the other hand these types of pose information are inserted at different locations into the network.
While the fine-grained joint locations are provided to the network as input, the view information is used to fuse the final CNN feature layers. 
Thus the combination of both extensions provides the network with pose information throughout a larger part of the network.

Figure \ref{fig:pseModel} shows the combination of the views model and the pose model.
Because of their modularity, both extensions can easily be used together as a combined extension to the baseline CNN model.
The first layer is adapted to be fed with the input image and the pose maps from the pose estimator like in the pose model of Section \ref{section:pse:pose}.
Additionally, and the \viewpredictor branch is added to estimate the view angle of the input image and fuse the \viewunits results accordingly. 
As the \viewpredictor is branched off the common CNN trunk fed with the input image and the pose maps, pose maps can also help to improve view estimation.

\begin{figure}
\centering
\includegraphics[width=\columnwidth]{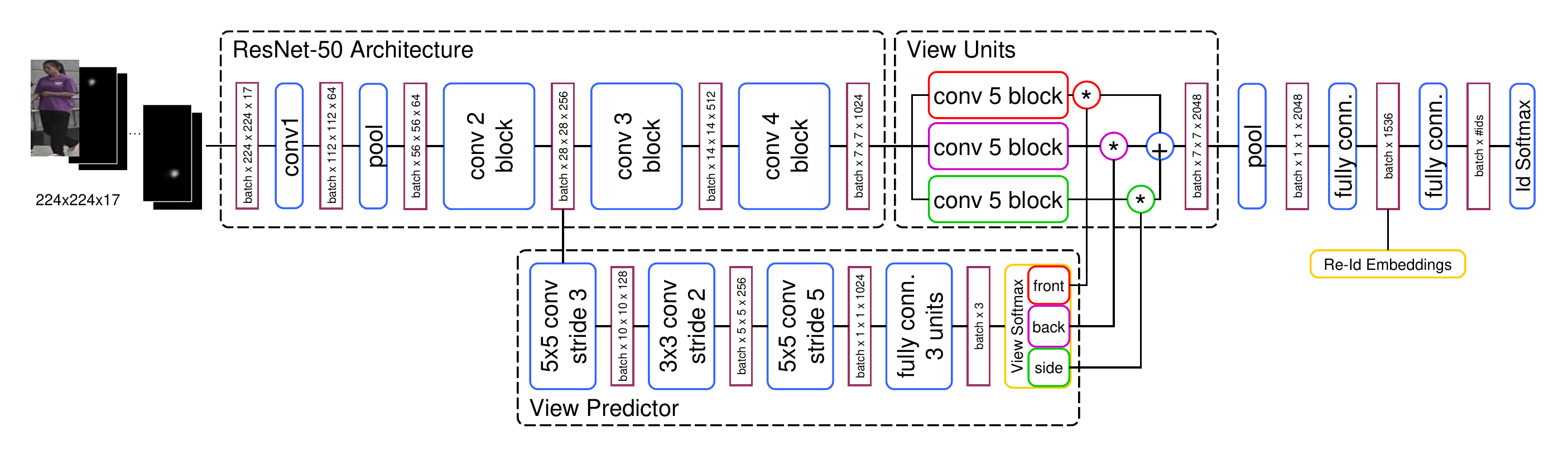}
\caption{PSE model based on ResNet-50. Blue boxes are network blocks like convolutions, pooling layers or complete ResNet blocks. Purple blocks between network blocks give the dimensions of the vector passing from one block to the next.}
\label{fig:pseResNetModel}
\end{figure}

Figure~\ref{fig:pseResNetModel} illustrates the PSE model based on the ResNet-50 architecture \cite{ResNet}. 
On the left side, the input image (3 channels) and the pose maps (14 channels) are fed into the network. 
After passing the standard ResNet-50 architecture's first convolutional and pooling layer, as well as the convolutional block 2, the \viewpredictor is branched off. 
In the main branch the network follows the standard ResNet architecture with convolutional blocks 3 and 4.
Convolutional block 5, however, is replicated three times, each of them forming a single \viewunit. 
The \viewunits' feature maps are then multiplied with the predicted view values and added to form the final convolutional feature map, which is then, again like in the standard ResNet-50, pooled.
The pooled features are then passed through a fully connected layer to form an embedding of size 1536. 
For training, another fully connected layer is applied to the embeddings and learned with a standard classification loss.

The \viewpredictor branch applies three convolutions to reduce the spatial dimensions to one. 
To achieve this, the first convolution applies a stride of three and the second a stride of two, both using padding. 
The third convolution with a kernel size of 5x5 is then applied without padding reducing the spatial dimension from five to one.
At the end, a fully connected layer is used to create the final view predictions.
Usage of the \emph{softmax} layer ensures a sum of one for the three view prediction values and therefore a normalized weighting of the \viewunits' feature maps.

\section{Training Details}
In this section, the details of the training procedure are described. 
While Section~\ref{section:pse:training:general} details the general training procedure, Section~\ref{section:pse:training:preTrainView} and Section~\ref{section:pse:training:pose} provide insight into the specialties of training the views and pose extensions.

\subsection{General Training Procedure}
\label{section:pse:training:general}
The training of the pose and view models are started by initializing them with an ImageNet pre-trained model.
Layers with changed dimensions (\eg the final classification layer) or newly added layers of the proposed advanced models, are randomly initialized. 

During the first training step, only these newly initialized layers are trained while the other layers remain fixed. 
This helps to prevent a negative impact of the randomly initialized layers on the well-trained layers that could otherwise negate the advantages of using a pre-trained model. 
Training of the first step is finished when the loss stagnates.
In the second step, the whole network is trained until the loss converges.

Training is performed with an Adam optimizer at recommended parameters. Training is started with a learning rate of $0.0001$ and a learning rate decay of $0.96$ is applied after each epoch.

In order to introduce more variance into the data and thus improve learning, basic data augmentation is applied to training images.
This is done by first resizing the image to 105\% width and 110\% height before randomly cropping it to the network's standard input size (224 by 224 pixels for Resnet-50 and 299 by 299 pixels for Inception-v4).
Furthermore, random horizontal flips are applied.

\subsection{Pre-Learning View Information}
\label{section:pse:training:preTrainView}
In general, it cannot be assumed that view information is available for the training data. For example, the widely used Market-1501 \cite{Market1501Dataset}, MARS \cite{MarsDataset} and DukeMTMC-reID \cite{DukeDataset} datasets for person re-id do not contain any information on the person's angle towards the camera.
To still be able to utilize view information, the model's view predictor is pretrained on the RAP dataset \cite{RapDataset}.

Although the RAP dataset does contain labels for \back, \front, \right and \left, the classes \right and \left are combined to one \side class serving two purposes:
At first, random horizontal flips are applied for data augmentation (see Section \ref{section:pse:training:general}), which result in \left and \right images being converted into each other. 
Thus the features for \left and \right images will be very similar and it is left to the network to detect both views similarly. 
On the other hand, the combined \side class has about the same size as the \front and \back classes of RAP with \left and \right almost equally making up for half of it.
Therefore, using the three classes (\front, \back, \side) ensures training with equally sized classes.

\begin{figure}
\centering
\includegraphics[width=\columnwidth]{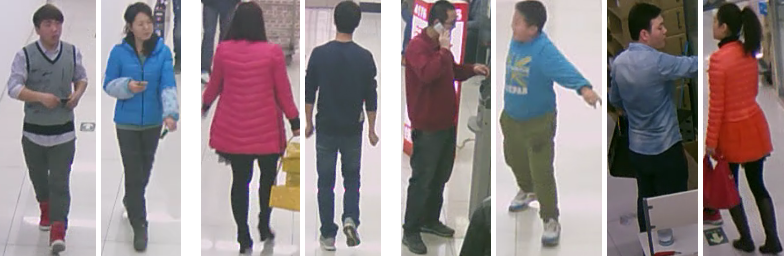}
\caption{Example images from the RAP dataset showing two \front, two \back and four \side view images.}
\label{fig:rapExamples}
\end{figure}

Figure \ref{fig:rapExamples} presents exemplary images from the RAP dataset.
The first two images are of the \front class, the next two of the \back class and the remaining four are from the \left and \right classes, which are combined into the \side class.
Note that while the \front and \back classes contain only images of a very narrow view angle range, the \side class contains images with a much larger view angle range resulting in a much larger variety in the \side class images.

To train the view model, at first, the model's view predictor side-branch is trained on the RAP dataset while the layers of the main network remain fixed. 
Following the proposed standard procedure, the network is initialized with an ImageNet pre-trained model and the view predictor's layers are randomly initialized.
Apart from this and that no pose maps are used, the procedure follows the same steps as shown in Figure~\ref{figure:pseTraining} on page~\pageref{figure:pseTraining}.
After the view predictor has been trained, the actual training for the target dataset is started by first training the newly initialized layers (\viewunits and \logits layers) while leaving the other layers fixed. 
When the loss has saturated, the complete network is trained until convergence.

\subsection{Training the Full Body Pose Model}
\label{section:pse:training:pose}

\begin{figure}
\centering
\includegraphics[width=.7\linewidth]{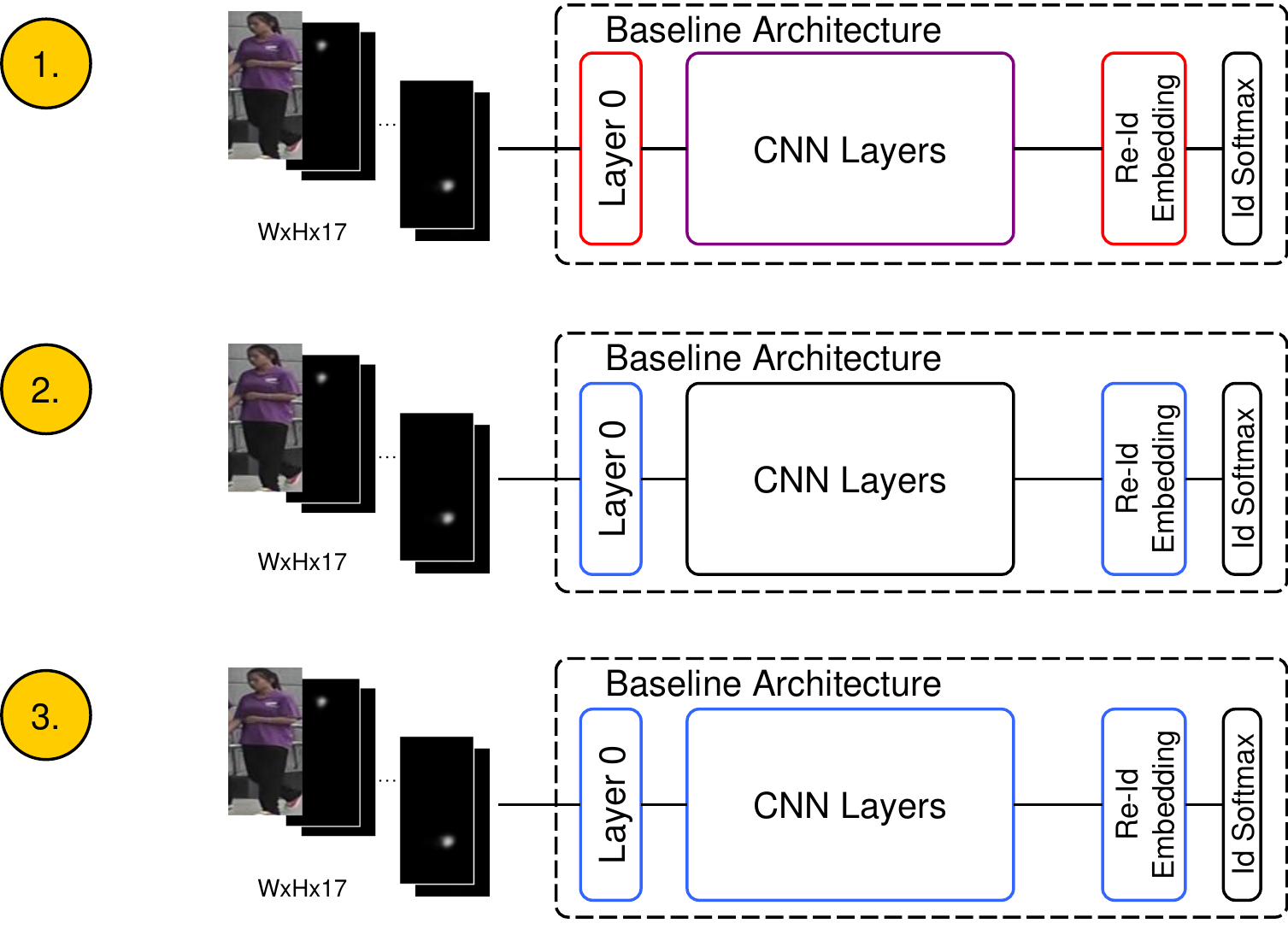}
\caption{Visualization of training steps for the full body pose model. In the first step, the model is initialized from a pre-trained ImageNet model where possible (purple layers). Layers with different dimensions (marked red) are randomly initialized. In the second step, the newly initialized layers are trained. Afterwards, the whole model is trained.}
\label{figure:poseModelTraining}
\end{figure}

As described in Section \ref{section:pse:pose}, the full body pose information is provided to the network by adding 14 additional input channels.
Hence, the first layer of the network cannot be initialized from the pre-trained ImageNet model but needs to be randomly initialized.

Figure~\ref{figure:poseModelTraining} visualizes the training procedure for the pose model.
In the first step, all layers of the baseline CNN architecture that have not changed in dimensions (marked purple) are initialized from the ImageNet pre-trained model.
The first and last layer of the network are randomly initialized (marked in red).

In step two and three, the proposed general training procedure is used by first training all randomly initialized layers (marked in blue) and keeping the pre-trained layers fixed to prevent damaging the well-formed layers.
Afterwards, the whole network is trained until convergence.

\subsection{Full PSE Training Procedure}
\label{section:pse:training:pse}

\begin{figure}
\centering
\includegraphics[width=\linewidth]{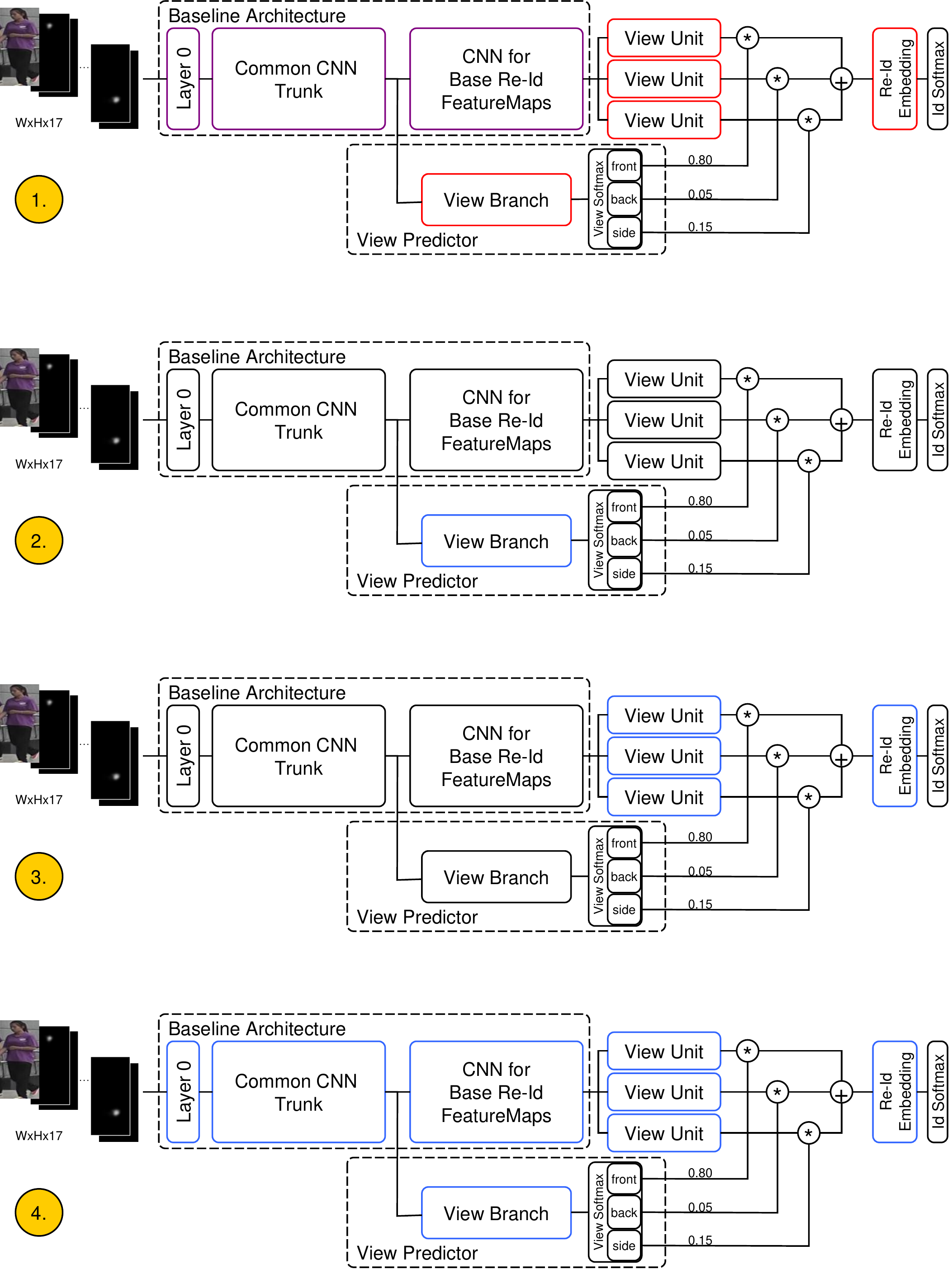}
\caption{Visualization of the four training steps for the proposed PSE model. In the first step the model's unchanged layers are initialized from a pre-trained pose model (purple) and randomly initialized for added layers (red). During further steps, the layers marked blue are trained while the others remain fixed.}
\label{figure:pseTraining}
\end{figure}

Training of the PSE model is done in multiple steps.
Because the PSE model is fed with the 17 channel-wide input consisting of the input image and the 14 pose maps channels, training is not directly initialized from an ImageNet pre-trained model.
Instead, the PSE model is initialized with the full body pose model trained as described in Section~\ref{section:pse:training:pose}.

Figure~\ref{figure:pseTraining} visualizes the four steps used to train the PSE model.
In the first step, all layers in the `Baseline Architecture' part of the model (marked in purple), are initialized from a pre-trained full body pose model. 
The layers marked with red, are freshly initialized from random as they either do not appear in the full body pose model (the \viewpredictor), have changed dimensions (the re-id embedding layer) or have been replicated (the \viewunits).

In the second step, the RAP dataset is used since it provides view labels to train the \viewpredictor side-branch (marked in blue). 
All other layers are fixed for this training step.
This ensures that training of the \viewpredictor does not alter the common network trunk which the latter layers of the main CNN branch are based and depend on.
The second step is only needed because the target datasets (\ie Market-1501, Duke and MARS) do not contain view labels.

During the third step, the \viewunits and the re-id embedding layer (marked in blue) are trained, while keeping the rest of the network fixed.
This is done to pre-train the randomly initialized layers before training the full model.
This step is important to preventto prevent the randomly initialized layers damaging the pre-learned layers since their random initialization would result in random results during first training steps and thus in unfortunate gradients being back-propagated through the whole network.
In this and the next step, training is done on the target dataset (\eg Market-1501 or Duke).

The full PSE model is then trained until convergence in the fourth training step.

%% file: content/evaluation.tex

\chapter{Evaluation}
\label{ch:Evaluation}

In this chapter performance of the proposed Pose-Sensitive Embedding (PSE) will be evaluated on two popular baseline CNN architectures: ResNet-50 \cite{ResNet} and Inception-v4 \cite{InceptionV4}.
Extensive experiments with various datasets show the generality of this approach.

Results are reported using the standard cross camera evaluation in the single query setting.
Performance is measured in rank scores calculated from cumulative matching characteristics (CMC) and mean average precision (mAP) in percent.
A CMC rank-$x$ score gives the averaged probability of a correct probe being part of the first $x$ retrieved gallery images.

The mAP provides a measure of quality across the recall levels and has shown to have especially good stability and discrimination.
With a query $q_j \in Q$ from the set of query images, let $\{d_1, \cdots, d_{m_j}\}$ be the relevant gallery images and $R_{jk}$ the set of ranked retrieval results from the top until gallery image $d_k$ is found.
Then, Equation~\ref{equation:map} gives the mAP.

\begin{equation}
mAP(Q)=\frac{1}{|Q|} \sum_{j=1}^{|Q|} \frac{1}{m_j} \sum_{k=1}^{m_j}\texttt{Precision}(R_{jk})
\label{equation:map}
\end{equation}

The methods proposed in this work are evaluated on three popular datasets.
Table~\ref{table:datasets} gives an overview over these datasets' characteristics.

\begin{table*}
  	\centering
        \begin{tabular}{ | l || C{3cm} | C{3cm} | C{3cm} |}
            \hline
                              	& Market-1501 	& Duke-MTMC-reID 	& MARS  	\\
            \hline
            \hline
            \# training ids 	& 751 			& 702 				& 625 	 	\\
            \# training images 	& 12,936 		& 16,522 			& 509,914 	\\
            \hline
            \# test ids 		& 750 			& 702 				& 636 		\\
            \# test images 		& 19,732 		& 17,661 			& 681,089 	\\
            \hline
            \# query images 	& 3,368 		& 2,228 			& 114,493 	\\
            \hline
            \# cameras 			& 6 			& 8 				& 6 		\\
            \hline
        \end{tabular}
    \caption{Comparison of the three datasets Market-1501 \cite{Market1501Dataset}, Duke-MTMC-reID \cite{DukeDataset} and MARS \cite{MarsDataset} used during the main evaluation.}
    \label{table:datasets}
\end{table*}

\begin{figure}
\centering
\begin{minipage}{.5\textwidth}
  \centering
  \includegraphics[height=3.3cm]{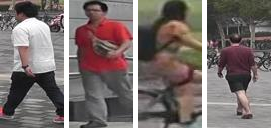}
  \captionof{figure}{Example images of the Market dataset.}
  \label{fig:market_examples}
\end{minipage}%
\begin{minipage}{.5\textwidth}
  \centering
  \includegraphics[height=3.3cm]{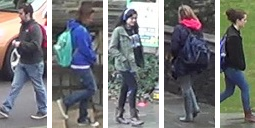}
  \captionof{figure}{Example images of the Duke dataset.}
  \label{fig:duke_examples}
\end{minipage}
\end{figure}

The Market-1501 (Market) dataset \cite{Market1501Dataset} consists of 32,668 annotated bounding boxes of 1,501 distinct persons. 
The bounding boxes were generated by a DPM person detector on videos from six cameras with non-overlapping views and differing quality.
For training 751 and for testing 750 persons are used.
The training, test and query set contain 12,936, 19,732 and 3,368 images, respectively.
All images have an equal height of 128 pixels and a width of 64 pixels.
Thus, all the person image's aspect ratios are changed equally when scaling them for detection via the used networks.
Figure~\ref{fig:market_examples} shows exemplary images of the Market dataset.

The Duke-MTMC-reID (Duke) dataset \cite{DukeDataset} is a subset of the DukeMTMC~\cite{DukeMTMC} dataset.
Person images are taken every 120 frames from 85-minute long high-resolution videos of eight cameras. 
Bounding boxes are hand-drawn.
All in all, the dataset contains 36,411 person images of 1,404 ids appearing in at least two cameras and 408 ids appearing in only one camera. 
The latter are added to the gallery set as distractor ids.
Image's width and height as well as aspect ratio are varying largely over the dataset. 
Thus their aspect ratio will be changed differently when resized to a fixed height and width.
Examples for the Duke dataset can be seen in Figure~\ref{fig:duke_examples}

\begin{figure}
\centering
\includegraphics[width=\columnwidth]{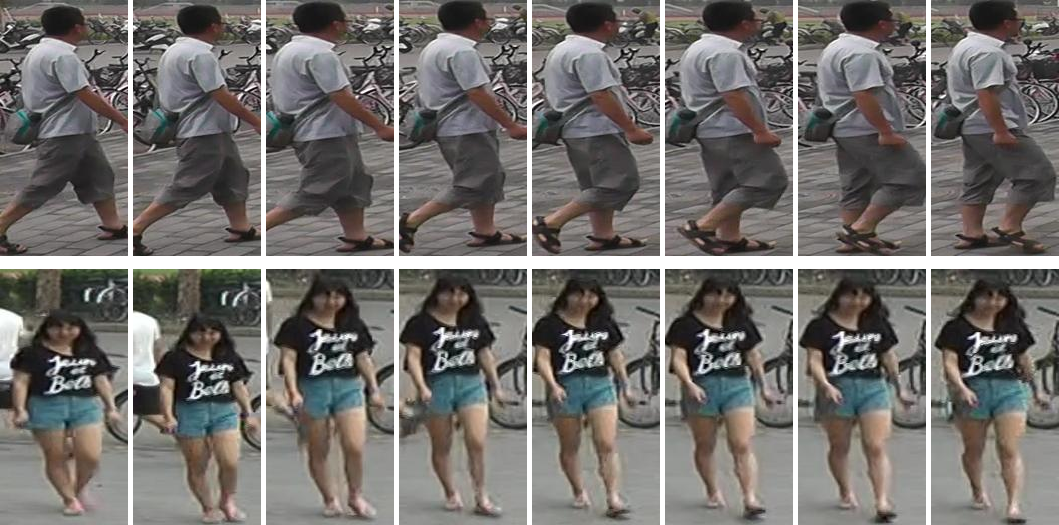}
\caption{Images from two tracklets of the MARS dataset.}
\label{fig:marsTrackletsExamples}
\end{figure}

The MARS dataset \cite{MarsDataset} is based on the same raw data as the Market dataset and the same persons have been assigned the same labels. 
In contrast to Market, MARS is providing tracklets of persons instead of single images. 
Therefore MARS is well suited to evaluate the performance of re-id approaches for person track retrieval. 
The dataset consists of 8,298 tracklets for training and 12,180 tracklets for testing with 509,914 and 681,089 images respectively.
In Figure~\ref{fig:marsTrackletsExamples} images from two tracklets of the MARS dataset are shown.

The evaluation is split into three sections. 
In Section~\ref{section:studyOfPose} the main evaluation is focusing on investigating the effects of including pose information and the comparison with the state of the art. 
The X-MARS reordering of the MARS dataset is introduced and the evaluation of image re-id systems on video data are discussed in Section~\ref{section:xmars}.
Finally, in Section~\ref{section:furtherRealWorldAspects} further real world aspects like the performance impact of large gallery sizes and automated person detection are investigated.

\section{Study of Pose Information}
\label{section:studyOfPose}
In this Section various aspects of introducing view information of different kinds into the network are evaluated and a comprehensive comparison with the state of the art is given.
At first, in Section~\ref{section:evaluation:viewVsPoseVsBoth} the effects of introducing either view or pose or using both are investigated. 
Section~\ref{section:evaluation:viewEstimation} researches the quality of the views prediction learned on RAP and applied on Market and Duke. 
In Section~\ref{section:evaluation:PSEvsPIE} the proposed method of directly providing the network with pose maps is compared to the explicit pose alignment done by Pose Invariant Embedding \cite{poseInvariantEmbedding}. Finally, in Section~\ref{section:stateOfTheArt} PSE is compared with the current state of the art, with and without re-ranking.

\subsection{View vs. Pose vs. Both}
\label{section:evaluation:viewVsPoseVsBoth}
To evaluate the usefulness of including different pose information, separate experiments including only coarse view information, fine-grained joint locations and the combination of both are conducted. 
These experiments are performed across the Market and Duke datasets.
Furthermore, to show the generality of the approach towards the underlying CNN architecture, experiments are performed with the main ResNet-50 and the popular Inception-v4 CNN architectures. 
For Inception-v4, the \viewpredictor is branched of the main model after the Reduction-A block.
Similarly, \viewunits are added by replacing the last Inception-C block with three parallel Inception-C blocks at the end. 
Results of these experiments are presented in Table~\ref{table:ablation_pose}.

\begin{table*}
\centering
\resizebox{\linewidth}{!}{
\begin{tabular}{ | l | l | ccccc | ccccc |} 
    \hline
	CNN & Method & \multicolumn{5}{c|}{Market-1501} & \multicolumn{5}{c|}{Duke}  \\
	                        & & mAP 		& R-1 		& R-5 		& R-10 		& R-50  	& mAP & R-1 & R-5 & R-10 & R-50 \\
	\hline\hline
    Inception-v4 & Baseline   & 51.9 		& 75.9 		&  89.8		& 92.5 		& 97.3 		& 36.6 &     61.8 & 74.8&	79.8	& 89.4  \\
	             & Views only & 61.9 		& 81.5		&  92.3		& 94.9		& 98.1 		& 40.3 &     62.7   & 76.6&	81.1	& 90.3 \\
		         & Pose only  & 60.9 		& 81.7		&  91.8		& 94.4 		&	97.9  	& 48.2 &     70.5  & 81.9 &	86.1&	92.7   \\
                 & PSE        & \f{64.9}	& \f{84.4}	& \f{93.1}	& \f{95.2} 	& \f{98.4} 	& \f{50.4} & \f{71.7}   & \f{83.5}	& \f{87.1} &	\f{93.1} \\
	\hline
	ResNet-50    & Baseline   & 59.8 		& 82.6 		& 92.4		& 94.9 		& 98.2  	& 50.3 &     71.5   &83.1&	87.0	& 94.1  \\
           		 & Views only & 66.9 		& \f{88.2} 	& \f{ 95.4}	& \f{97.2}	& 98.9  	& 56.7 &     76.9  &  87.3&	90.7	& 95.7    \\
		   		 & Pose only  & 61.6 		& 82.8 		& 93.1		& 95.5		& 98.3  	&      53.1 &     73.4  &  84.5	&88.1 &	94.3  \\
           		 & PSE		  & \f{69.0} 	& 87.7 		& 94.5		& 96.8 		& \f{99.0}  &    \f{62.0} & \f{79.8}  &  \f{89.7} & \f{92.2} & \f{96.3}\\
	\hline
\end{tabular}
}
\caption{Evaluation of the effects of introducing different kinds of pose information. While coarse view information and fine-grained joint locations each lead to notable gains, the combination of both yields further improvements most of the times.}
\label{table:ablation_pose}
\end{table*}

In comparison to the baseline models without any explicitly modeled pose information, inclusion of either views or pose yield significant improvements in mAP and rank scores. 
This can be observed across both datasets and both underlying CNN architectures. 
Inclusion of views into the ResNet-50 model provides consistent improvements of $7.1\%$ mAP and $5.6\%$ rank-1 on Market as well as $6.4\%$ mAP and $5.4\%$ rank-1 for Duke.
Even though the absolute improvements are smaller when pose information is used, they still surpass the baseline model by about $2$-$3\%$ in mAP.
While both, pose and view information yield improvements for the Inception-v4 architecture, results are less consistent. 
For Market, pose and view achieve almost equal improvements of $9-10\%$ in mAP and about $6\%$ in rank-1.
In contrast, for Duke, pose information provides a much higher boost of $11.6\%$ mAP and $8.7\%$ rank-1 compared to $3.7\%$ mAP and $0.9\%$ rank-1 with view information.

\begin{figure}
\centering
\includegraphics[width=\columnwidth]{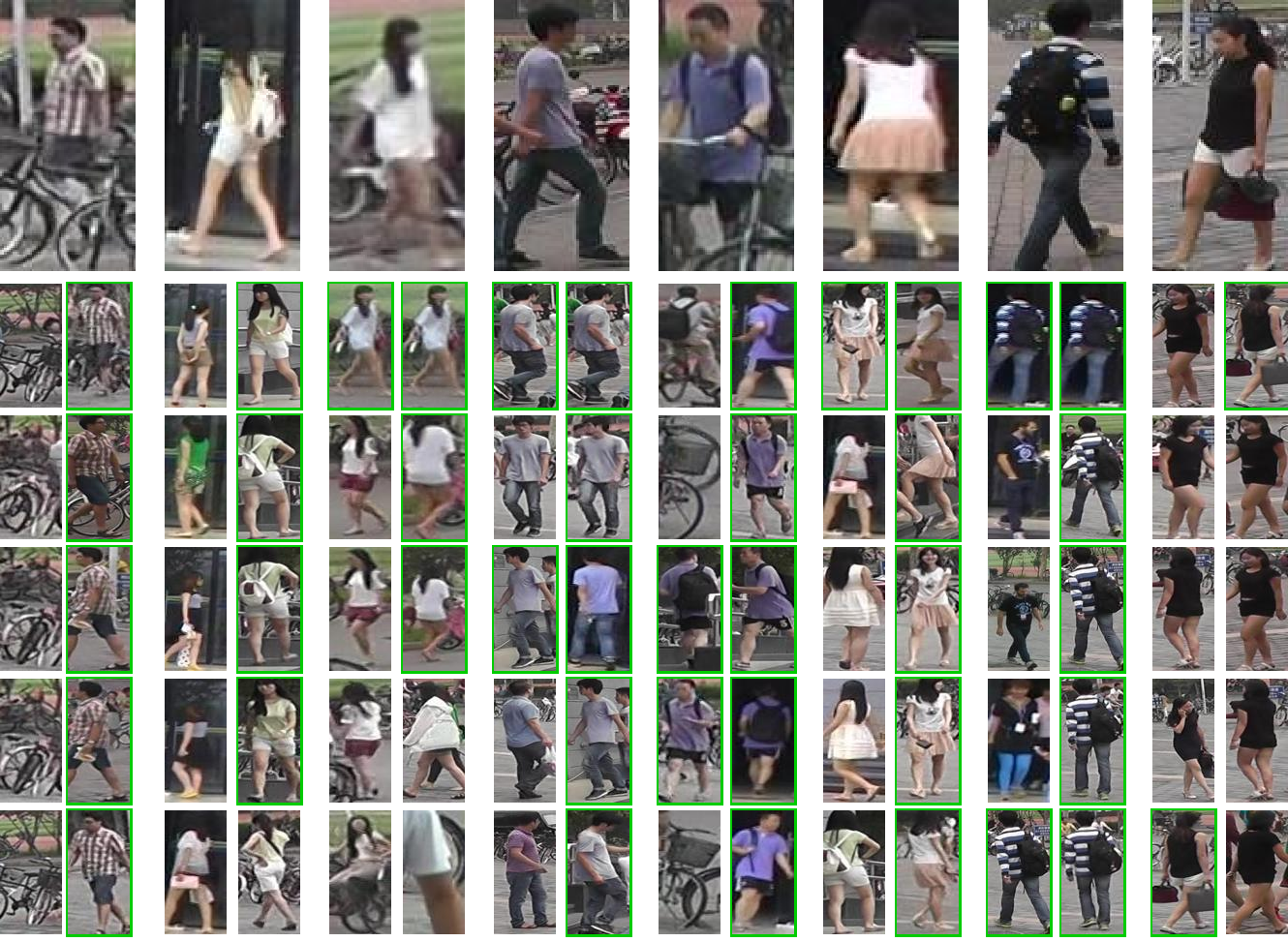}
\caption{Qualitative examples from the Market dataset to illustrate improvements of PSE over the baseline model. Below the query image shown on top, the left and right columns show the top five retrieved gallery images for the baseline and the PSE model. Correct retrievals are marked with a green border.}
\label{figure:queriesAndGalleries}
\end{figure}

Finally, the combination of both types of view information leads to a further consistent increase in mAP. 
With the ResNet-50 model mAP is increased by $2.1\%$ on Market and $5.3\%$ on Duke in comparison to the best result of either views or pose.
Similarly, for Inception-v4, mAP is improved by $3.0\%$ on Market and $2.2\%$ on Duke respectively. 
These results clearly show that the proposed method to include different types of view information benefits person re-id and indicates they complement each other.

Figure~\ref{figure:queriesAndGalleries} shows the retrieved gallery images for several exemplary query images from the Market dataset.
The top images display the query images used for matching.
The left and right columns below each query image show the top five retrieved images for the baseline and the PSE model respectively. 
Matching images are marked with a green border.
This qualitative overview gives a good impression of the improvements gained with the PSE model which is retrieving fewer distractor images and more matching images with different view angles than the query.

\subsection{Study of View Estimation}
\label{section:evaluation:viewEstimation}
\begin{figure}[b]
\centering
\includegraphics[width=\columnwidth]{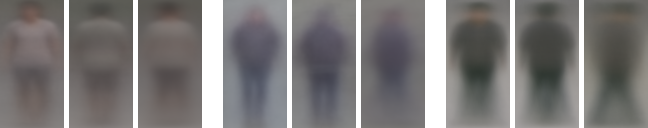}
\caption{Mean images of Market-1501 (left), Duke (center) and RAP (right) test sets using predictions of the PSE model's view predictor. The images show the pixel-wise average over \front, \back and \side view images from left to right.}
\label{figure:meanImages}
\end{figure}

Due to the fact that the evaluated person re-id datasets do not provide view angle information, the models' \viewpredictor is pre-learned on the RAP dataset as described in Section~\ref{section:pse:training:preTrainView}.
After training the ResNet-50 \viewpredictor on the RAP training set, it achieves an accuracy of $82.2\%$, $86.9\%$ and $81.9\%$ on the annotated RAP test set for \front, \back and \side views, respectively.
As no view labels are available on Market and Duke, the accuracy of the view prediction for these datasets cannot be calculated.
To still give a qualitative insight into the view prediction, Figure~\ref{figure:meanImages} shows mean images for Market, Duke and RAP.
These mean images are calculated for each dataset by averaging all test set images in their respective view classes, as estimated by the \viewpredictor.

For all three datasets, the mean images for \front and \back views are clearly distinguishable.
While the lower image part with the legs looks quite similar between \front and \back, the upper part of the image and especially the head region shows a difference between the views. 

In comparison, the \side view mean images look more like an in-between of \front and \back images with very blurry leg regions.
The increased variance in the \side view mean images has multiple sources.
At first, the \side view class contains images with persons turned to the right and to the left resulting in a mix of right and left images.
Secondly, in the RAP dataset all images not being totally frontal or backwards are labeled with the side view classes (\left and \right).
Thus, the images in the side view classes already have a much larger variance in the training dataset.
At last, the leg region of the images is more blurry, as walking people's legs have varying positions depending on when the image is taken during their steps.

\begin{figure}[b!]
\centering
\includegraphics[width=0.7\columnwidth]{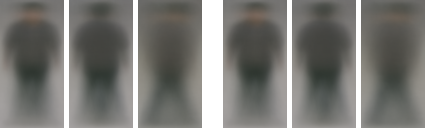}
\caption{Comparison of mean view images of RAP test set. On the left side mean images of \front, \back and \side images estimated by the \viewpredictor are shown. On the right side, mean images have been calculated by using the annotated views.}
\label{figure:rapMeanImages}
\end{figure}

Additionally, in Figure~\ref{figure:rapMeanImages}, the mean images of the RAP test set are compared by using the predicted view for the left and by using the annotated view for the right images. 
Although RAP test accuracy is between $81.9\%$ and $86.9\%$, the mean images look very similar. 
This indicates that mis-detections happen for very similar images at the border between view classes, thus not leading to large changes in the resulting mean image.
Furthermore, the same effect of a much blurrier \side image is observed in the annotated case, confirming that this effect is based on the RAP dataset's view labeling.

All in all, comparison of the mean images gives a qualitative indication that the transfer of view prediction from RAP to Market and Duke works well. 
This is especially interesting as all three datasets have been recorded in varying conditions (indoor \vs outdoor; warm \vs cold temperatures) leading to largely differing clothing of the recorded persons.

\subsection{Explicit use of Pose Information}
\label{section:evaluation:PSEvsPIE}

Pose Invariant Embedding (PIE) \cite{poseInvariantEmbedding} is another approach utilizing pose information to improve person re-id performance. 
In contrast to the PSE method, PIE uses the estimated pose information to explicitly align body parts by generating a \emph{PoseBox} image.
The network is then fed with the original image, the PoseBox image and the pose estimation's confidence score.
A further difference is that for PIE the base CNN model's convolutional layers are duplicated with one branch being fed the original person image and the other branch being fed with the PoseBox image. 
However, the duplication of all the CNN layers of the underlying architecture results in a large increase in parameter counts and thus impacts running times.
After the convolutional layers the feature maps of both branches are fused with the confidence score to create the final embedding.

\begin{table*}[b!]
\centering
\begin{tabular}{ | l | l | cccc |} 
    \hline
	\multicolumn{2}{|c|}{Method}  &    mAP & R-1 & R-5 & R-10  \\
	\hline\hline
   	\multirow{3}{*}{\rotatebox{90}{\hspace{-.2cm}PIE}}
    & Baseline1 (R,Pool5) 		& 47.6 		& 73.0 		& 87.4 		& 91.2 \\
    & PIE (R,Pool5) 			& 53.9 		& 78.7 		& 90.3 		& 93.6 \\
    \cline{2-6}
    & Difference to Baseline	& 6.3 		& \f{5.7}	& \f{2.9}	& \f{2.4} \\
    \hline
    \hline
   	\multirow{3}{*}{\rotatebox{90}{\hspace{-.2cm}Our}}
    & Baseline (ResNet-50) 		& 59.8 		& 82.6 		& 92.4 		& 94.9 \\
    & PSE (ResNet-50) 			& 69.0 		& 87.7 		& 94.5 		& 96.8 \\
    \cline{2-6}
    & Difference to Baseline	& \f{9.2} 	& 5.1		& 2.1		& 1.9 \\
    \hline
\end{tabular}
\caption{Comparison between best results for Pose Invariant Embedding (PIE) \cite{poseInvariantEmbedding} and Pose Sensitive Embedding (PSE). Both are compared using ResNet-50 as base CNN model and evaluation is done with the Market-1501 dataset.}
\label{table:poseMapsVsPIE}
\end{table*}

Table~\ref{table:poseMapsVsPIE} shows a comparison between the best results of PIE and PSE on the Market dataset, both using a ResNet-50 architecture as base model.
The PSE model clearly exceeds the PIE model by $15.1\%$ in mAP and $9.0\%$ in rank-1.
Even when comparing the absolute increase over the baseline model, the increase of PSE over its much stronger baseline is significantly larger in mAP with $9.2\%$ in comparison to $6.3\%$ mAP gained by PIE.

For rank accuracies the absolute improvement of PSE over its baseline is slightly worse than for PIE.
However, the absolute difference does not account for the effects of the much stronger baseline and hence the increase difficulty to achieve the same absolute improvements. 
When looking at the absolute reduction of error (\ie how much of the gap to $100\%$ is closed), PSE reduces the gap for rank-1 accuracies by $29.3\%$ whereas PIE only reduces it by $21.1\%$.

\subsection{State of the Art}
\label{section:stateOfTheArt}

In Table~\ref{table:SOTA} the state of the art is compared with the performance of the proposed Pose Sensitive Embedding on the three datasets Market, Duke and MARS.
In the upper section of the table, the proposed embedding is compared with published state of the art approaches without applying re-ranking.
On the MARS and Duke datasets, PSE based on a ResNet-50 baseline architecture achieves top accuracies.
On Market, it performs slightly worse than DPFL~\cite{DPFL}, which employs two or more multi-scale embeddings for retrieval, resulting in  a significantly larger workload during evaluation.
Across all three datasets, a consistent improvement over the ResNet-50 Baseline model between $7.4\%$ to $12.3\%$ in mAP and $5.1\%$ to $8.3\%$ in rank-1 is observed.

\begin{table*}[b!]
\centering
\resizebox{\linewidth}{!}{
\begin{tabular}{ | l | l | l | cc | cc | cc |}
    \hline
    \multicolumn{3}{|c|}{Method} & 
	\multicolumn{2}{c|}{Market-1501} & 
    \multicolumn{2}{c|}{Duke}  &
    \multicolumn{2}{c|}{MARS}  \\

    \multicolumn{3}{|c|}{}                  & mAP  & R-1  & mAP  & R-1  & mAP  & R-1  \\
	\hline\hline
    \mcl{P2S\cite{P2S}}                     & CVPR17        & 44.3 & 70.7 &    - &    - &    - &    - \\
    \mcl{Spindle\cite{zhao2017spindle}}     & CVPR17        &    - & 76.9 &    - &    - &    - &    - \\
    \mcl{Consistent Aware\cite{ConsAw}}     & CVPR17	    & 55.6 & 80.9 &    - &    - &    - &    - \\
    \mcl{GAN\cite{GAN}}                     & ICCV17    	& 56.2 & 78.1 & 47.1 & 67.7 &    - &    - \\
    \mcl{Latent Parts \cite{Li_2017_CVPR}}  & CVPR17        & 57.5 & 80.3 &    - &    - & 56.1 & 71.8 \\
    \mcl{ResNet+OIM \cite{OIM}}	            & CVPR17 		&    - & 82.1 &    - & 68.1 &    - &    - \\
    \mcl{ACRN\cite{ACRN}}                   & CVPR17-W 		& 62.6 & 83.6 & 52.0 & 72.6 &    - &    - \\
    \mcl{SVD \cite{SVD}}                    & ICCV17  		& 62.1 & 82.3 & 56.8 & 76.7 &    - &    - \\
    \mcl{Part Aligned \cite{zhao2017deeply}}& ICCV17        & 63.4 & 81.0 &    - &    - &    - &    - \\
    \mcl{PDC \cite{su2017pose}} 	        & ICCV17		& 63.4 & 84.1 &    - &    - &    - &    - \\
    \mcl{JLML \cite{JLML}}                  & IJCAI17       & 65.5 & 85.1 &    - &    - &    - &    - \\  
    \mcl{DPFL \cite{DPFL}}                  & ICCV17-W  &\f{72.6}&\f{88.6}& 60.6 & 79.2 &    - &    - \\
    \mcl{Forest \cite{Forest}}              & CVPR17    	&    - &    - &    - &    - & 50.7 & 70.6 \\
    \mcl{DGM+IDE \cite{DGM}}	            & ICCV17 		&    - &    - &    - &    - & 46.8 & 65.2 \\
    \mcl{QMA \cite{QMA}}                    & CVPR17		&    - &    - &    - &    - & 51.7 & \f{73.7} \\
	\hline
    \multirow{2}{*}{\rotatebox{90}{\hspace{-.2cm}Our}}
    & \mcl{ResNet-50 Baseline}                              & 59.8 & 82.6 & 50.3 & 71.5 & 49.5 & 64.5 \\
    & \mcl{\textbf{PSE}}   	                                &69.0&87.7&\f{62.0}&\f{79.8}&\f{56.9}& 72.1 \\

    \hline\hline
    \mcl{Smoothed Manif. \cite{Bai_2017_CVPR}}          & CVPR17      & 68.8 & 82.2 &    - &    - &    - &    - \\
    \mcl{IDE (R)+XQDA + k-reciprocal \cite{zhong2017re}}  & CVPR17      & 61.9 & 75.1 &    - &    - & 68.5 & 73.9 \\ 
    \mcl{IDE (R)+KISSME + k-reciprocal \cite{zhong2017re}}& CVPR17      & 63.6 & 77.1 &    - &    - & 67.3 & 72.3 \\ 
    \mcl{DaF \cite{yu2017divide}}                       & BMVC17      & 72.4 & 82.3 &    - &    - &    - &    - \\
    \hline
    \multirow{2}{*}{\rotatebox{90}{\hspace{-.2cm}Our}}
    & \multicolumn{2}{|l|}{PSE + k-reciprocal \cite{zhong2017re}}      & 83.5 & 90.2 & 78.9 & 84.4 &    70.7 &	74.9   \\
   	& \multicolumn{2}{|l|}{\textbf{PSE + ECN (rank-dist) \cite{pse-ecn}}}  &\f{84.0}&\f{90.3}&\f{79.8}&\f{85.2} & \f{71.8} & \f{76.7} \\
	\hline
\end{tabular}
}
\caption{Comparison of the proposed PSE approach with the published state of the art. In the top section of the table, the PSE embedding is compared to state of the art methods not using re-ranking. In the lower part, re-ranked results are compared with re-ranked versions of PSE.}
\label{table:SOTA}
\end{table*}

In the lower section of Table~\ref{table:SOTA} re-ranked results of the PSE embedding are compared with state of the art re-ranked methods.
For re-ranking the PSE embedding, both, k-reciprocal embedding \cite{zhong2017re} and the new expanded cross neighborhood (ECN) re-ranking \cite{pse-ecn} are used.
On all datasets re-ranked PSE results achieve top accuracies with ECN being slightly better than k-reciprocal re-ranking. 
The state of the art is improved on Market by $11.6\%$ in mAP and $8.0\%$ in rank-1 and by $3.3\%$ in mAP and $8.0\%$ in rank-1 for MARS.
For Duke, no published re-ranked results have been found.

\section{X-MARS: Enabling Image to Video Evaluation}
\label{section:xmars}

With real-world applications of person re-id becoming more and more suitable, the amount of required data for training a re-id system becomes an important concern since labeling large amounts of images is a time-consuming and hence expensive task.
This holds especially true when working with video datasets to do re-id on tracklets instead of single images.
\Eg the MARS dataset providing tracklets is about 36 times larger than the Market dataset, while they are both based on the same data source and MARS even contains less identities than Market (see Table~\ref{table:datasets}).
In contrast, being able to train on a standard single-image person re-id dataset (like Market) and using that network for tracklet detection (as provided by MARS) would drastically reduce the number of images to be labeled.
Therefore, cross evaluating a person detection system trained on the Market dataset with the MARS dataset could give insights into these aspects.

In Section~\ref{section:xmars:xmars} it is discussed why the MARS dataset cannot be used for such an evaluation and a novel reordering of it called X-MARS is introduce to solve this shortcoming.
Subsequently, the proposed evaluation based on X-MARS is presented in Section~\ref{section:xmars:evaluation}.

\subsection{X-MARS}
\label{section:xmars:xmars}

Unfortunately, although the MARS and Market datasets are based on the same data source and labels are assigned consistently, they cannot be used for cross-evaluation since their test and training sets overlap largely (see Table~\ref{table:mars-x-mars-overlap}).
In fact, $48.3\%$ of the MARS test set identities are contained in the Market training set diminishing the significance of a cross evaluation.

\begin{table*}[b!]
  \centering
  \begin{tabular}{ | l  l | c  c |}
  	\hline
    & & \multicolumn{2}{c|}{Market-1501} \\
    & & train & test \\
	\hline
    \multirow{2}{*}{MARS}
    & train & 312& 307 \\
    & test & 313 & 329 \\
    \hline
  \end{tabular}
  \caption{Comparison of the overlap between the Market-1501 and MARS datasets. The table shows the number of identities shared between the respective dataset part of the MARS and Market-1501 datasets.}
  \label{table:mars-x-mars-overlap}
\end{table*}

To enable a meaningful cross-evaluation, a reordering of the MARS dataset's test and training splits called X-MARS is proposed.
Since the IDs in Market and MARS are consistent and the IDs used by MARS are a subset of the IDs used by Market, it is possible to reorder the tracklets of MARS based on the train/test split of Market. 
This is done by assigning all IDs of MARS (\ie the union of the test and training IDs) which are part of the Market training set to the X-MARS training set. 
The same procedure is applied for the test set of X-MARS.
In order to ease comparability between X-MARS and MARS and allow reusing the evaluation scripts for MARS with X-MARS, the query/gallery split and the used file format are created in the same way as it was done for MARS.

Table~\ref{table:x-mars-vs-mars} compares the MARS dataset and the reordered X-MARS.
While X-MARS has slightly less training identities than the original MARS dataset, the number of tracklets and images in the training and test sets do not differ significantly (\eg X-MARS has only $2.8\%$ less images in the training set).
The code for generating the training, query and gallery splits as well as the IDs of the splits and required files for evaluation are provided at \href{https://github.com/andreas-eberle/x-mars}{github.com/andreas-eberle/x-mars}.

\begin{table*}
  \centering
  \resizebox{\linewidth}{!}{
  \begin{tabular}{ | l | c | c | c | c | c | c | c | c | c | c |}
  	\hline
    & \multicolumn{3}{c|}{Train} & \multicolumn{3}{c|}{Test} & \multicolumn{2}{c|}{Query} \\
    Dataset & \#IDs & \#Tracklets & \#Images & \#IDs & \#Tracklets & \#Images & \#Tracklets & \#Images \\
    \hline
    MARS & 625 & 8,298 & 509,914 & 636 & 12,180 & 681,089 & 1980 & 114,493 \\
    X-MARS & 619 & 7,986 & 495,857 & 642 & 12,492 & 695,146 & 2003 & 135,685 \\
    \hline
  \end{tabular}
  }
  \caption{Comparison of MARS and X-MARS datasets.}
  \label{table:x-mars-vs-mars}
\end{table*}

\subsection{X-MARS evaluation}
\label{section:xmars:evaluation}

In this section the performance of the proposed embeddings is discussed when trained on image data and evaluated on tracklets.
In Table~\ref{table:xMarsEvaluation} the performance of models trained on Market is compared when being evaluated on the Market and X-MARS test sets.
The results are shown for both of the baseline architectures, Inception-v4 and ResNet-50.

\begin{table*}
\centering
\resizebox{\linewidth}{!}{
\begin{tabular}{ | l | l | ccccc | ccccc |}
    \hline
    CNN & Method & \multicolumn{5}{c|}{Market-1501} & \multicolumn{5}{c|}{X-MARS} \\
    & & mAP & R-1 & R-5 & R-10 & R-50 & mAP & R-1 & R-5 & R-10 & R-50 \\
	\hline\hline
    Inception-v4 & Baseline & 51.9 		& 75.9 		&  89.8		& 92.5 		& 97.3 		& 50.5		& 70.6		& 82.2		& 85.1		& 91.9 \\
    & Views only 			& 61.9 		& 81.5		&  92.3		& 94.9		& 98.1 		& 58.5		& \f{76.0}	& 85.4		& 87.9		& 92.6 \\
    & Pose only 			& 60.9 		& 81.7		&  91.8		& 94.4 		&	97.9  	& 57.5		& 75.6		& 85.4		& 88.2		& 93.1 \\
    & PSE 					& \f{64.9}	& \f{84.4}	& \f{93.1}	& \f{95.2} 	& \f{98.4} 	& \f{58.7}	& 75.3		& \f{85.6}	& \f{88.6}	& \f{93.9} \\
    \hline
    ResNet-50 & Baseline 	& 59.8 		& 82.6 		& 92.4		& 94.9 		& 98.2  	& 49.9		& 69.8		& 80.3		& 83.9		& 90.1 \\
    & Views only 			& 66.9 		& \f{88.2} 	& \f{ 95.4}	& \f{97.2}	& 98.9  	& 56.4		& 73.2		& \f{85.2}	& 88.1		& 92.2 \\
    & Pose only 			& 61.6 		& 82.8 		& 93.1		& 95.5		& 98.3  	& 52.8		& 71.7		& 82.2		& 85.6		& 90.9 \\
    & PSE 					& \f{69.0} 	& 87.7 		& 94.5		& 96.8 		& \f{99.0}  & \f{59.2}	& \f{74.9}	& \f{85.2}	& \f{88.3}	& \f{92.9} \\
	\hline
\end{tabular}
}
\caption{Performance of the proposed embeddings trained on Market-1501 on the Market-1501 and the X-MARS test set to evaluate and compare transfer capabilities of the single-image learned models on tracklet data.}
\label{table:xMarsEvaluation}
\end{table*}

The final PSE embedding overall works best for X-MARS improving over $8.2\%$ and $9.3\%$ in mAP for Inception-v4 and ResNet-50 respectively.
Furthermore, it is again observed that while the views only and pose only models can largely improve re-id, the combination of both yields even better results in most of the cases.
The tendency that view information seems more important is also confirmed for the X-MARS evaluation.

An interesting observation is that the Inception-v4 model, while getting an about $3\%$-$8\%$ worse performance on Market than ResNet-50, is achieving about the same or even better performance than ResNet-50 on X-MARS. 
This holds true across all the models as well as the baseline, suggesting the Inception-v4 baseline model is able to generalize better for the X-MARS data in this scenario.

\section{Further Real World Aspects}
\label{section:furtherRealWorldAspects}

For real world applications of person re-id, further aspects need to be considered.
One such aspect is that automated detection systems need to be able to work with very large gallery sizes as images from many cameras need to be fed into the system.
Section~\ref{section:scalabiltyWithLargeGallerySizes} investigates the impact of large gallery sizes on the performance of the embedding.
Another important issue is the usage of automatically detected person images in comparison to using hand-drawn bounding boxes, since an automated detection system produces false detections and bad alignments. 
The effect of these errors on re-id is discussed in Section~\ref{section:prw}.

\subsection{Scalability with Large Gallery Sizes}
\label{section:scalabiltyWithLargeGallerySizes}

To evaluate the robustness in real-world deployments with very large gallery sizes, the scalability of the PSE model is investigated with the Market-1501+500k (Market500k) dataset.
This dataset is an extension to the Market dataset offering an additional 500,000 distractor images that can be added to the gallery to evaluate the impact of very large gallery sizes.
To examine the impact of added distractors, the accuracy of the proposed models is evaluated with increasing gallery sizes by adding 100k, 200k, 300k, 400k and 500k distractor images randomly sampled from the Market500k dataset's distractors.

\begin{table*}[b!]
\centering
\begin{tabular}{ | l|l  | cccc | cccc |}
    \hline
    \mcl{Method} & \multicolumn{4}{c|}{mAP by \#Distractors} & \multicolumn{4}{c|}{R-1 by \#Distractors} \\

    \mcl{}                          &    0 & 100k & 200k & 500k &    0 & 100k & 200k & 500k \\
	\hline\hline
    \mcl{I+V\np \cite{IV}} 			& 59.9 & 52.3 & 49.1 & 45.2 & 79.5 & 73.8 & 71.5 & 68.3 \\
    \mcl{APR\np$^*$ \cite{attid}} 	& 62.8 & 56.5 & 53.6 & 49.8 & 84.0 & 79.9 & 78.2 & 75.4 \\
    \mcl{TriNet\np$^\S$ \cite{trinet}} 	&\f{69.1}&61.9 & 58.7 & 53.6 & 84.9 & 79.7 & 77.9 & 74.7 \\
    \hline
    \multirow{4}{*}{\rotatebox{90}{\hspace{-.1cm}Our}}
    & ResNet-50 Baseline 	        & 59.8 & 54.6 & 51.8 & 47.5 & 82.6 & 77.7 & 75.7 & 73.2 \\
    & Views Only 			        & 66.9 & 61.5 & 58.9 & 54.8 & \f{88.2} & \f{84.4} & \f{83.2} & \f{81.2} \\
    & Pose Only 			        & 63.0 & 57.7 & 54.9 & 50.6 & 83.6 & 80.0 & 77.9 & 75.1 \\
    & PSE 			                & 69.0 & \f{63.4} & \f{60.8} & \f{56.5} & 87.7 & 84.1 & 82.6 & 80.8 \\
	\hline
\end{tabular}
\caption{Comparison of the view, pose and PSE models' embeddings on the Market-1501+500k distractors dataset ($\dag$ = unpublished works, $*$ = additional attribute ground truth, $\S$ = $\times 10$ test-time augmentation).}
\label{table:market500k}
\end{table*}

\begin{table*}[b!]
\centering
\resizebox{\linewidth}{!}{
\begin{tabular}{ | l|l  | cccc | cccc |}
    \hline
    \mcl{Method} & \multicolumn{4}{c|}{mAP by \#Distractors} & \multicolumn{4}{c|}{R-1 by \#Distractors} \\

    \mcl{}     		                    &    0 		& 100k 		& 200k 		& 500k 		& 0 	& 100k 		& 200k 		& 500k \\
	\hline\hline
    \mcl{I+V\np \cite{IV}} 				& 59.9 		& -12.7\% 	& -18.0\% 	& -24.5\% 	& 79.5 	& -7.2\% 	& -10.1\%	& -14.1\% \\
    \mcl{APR\np$^*$ \cite{attid}} 		& 62.8 		& -10.0\% 	& -14.7\% 	& -20,7\% 	& 84.0 	& -4.9\% 	& -6.9\% 	& -10.2\% \\
    \mcl{TriNet\np$^\S$ \cite{trinet}} 	& 69.1		& -10.4\% 	& -15.1\% 	& -22.4\% 	& 84.9 	& -6.1\% 	& -8.5\% 	& -12.0\% \\
    \hline
    \multirow{4}{*}{\rotatebox{90}{\hspace{-.1cm}Our}}
    & ResNet-50 Baseline 	        	& 59.8 		& -8.7\% 		& -13.4\% 		& -20.6\% 		& 82.6 	& -5.9\% 		& -8.4\% 		& -11.4\% \\
    & Views Only 			        	& 66.9 		& \f{-8.1\%} 	& -12.0\% 		& \f{-18.1\%} 	& 88.2 	& -4.3\% 		& \f{-5.7}\% 	& \f{-7.9}\% \\
    & Pose Only 			       		& 63.0 		& -8.4\% 		& -12.9\% 		& -19.7\% 		& 83.6 	& -4.3\% 		& -6.8\% 		& -10.7\% \\
    & PSE 			                	& 69.0 		& \f{-8.1\%} 	& \f{-11.9\%} 	& \f{-18.1}\% 	& 87.7 	& \f{-4.1}\% 	& -5.8\% 		& \f{-7.9}\% \\
	\hline
\end{tabular}
}
\caption{Evaluation of performance drop of the proposed embeddings and related works on the Market-1501+500k distractors dataset. The `0' columns state the detection accuracy without added distractors. The further columns give the accuracy decrease of the respective model. ($\dag$ = unpublished works, $*$ = additional attribute ground truth, $\S$ = $\times 10$ test-time augmentation).}
\label{table:market500kRelative}
\end{table*}

\begin{figure}
\centering
\includegraphics[width=\linewidth]{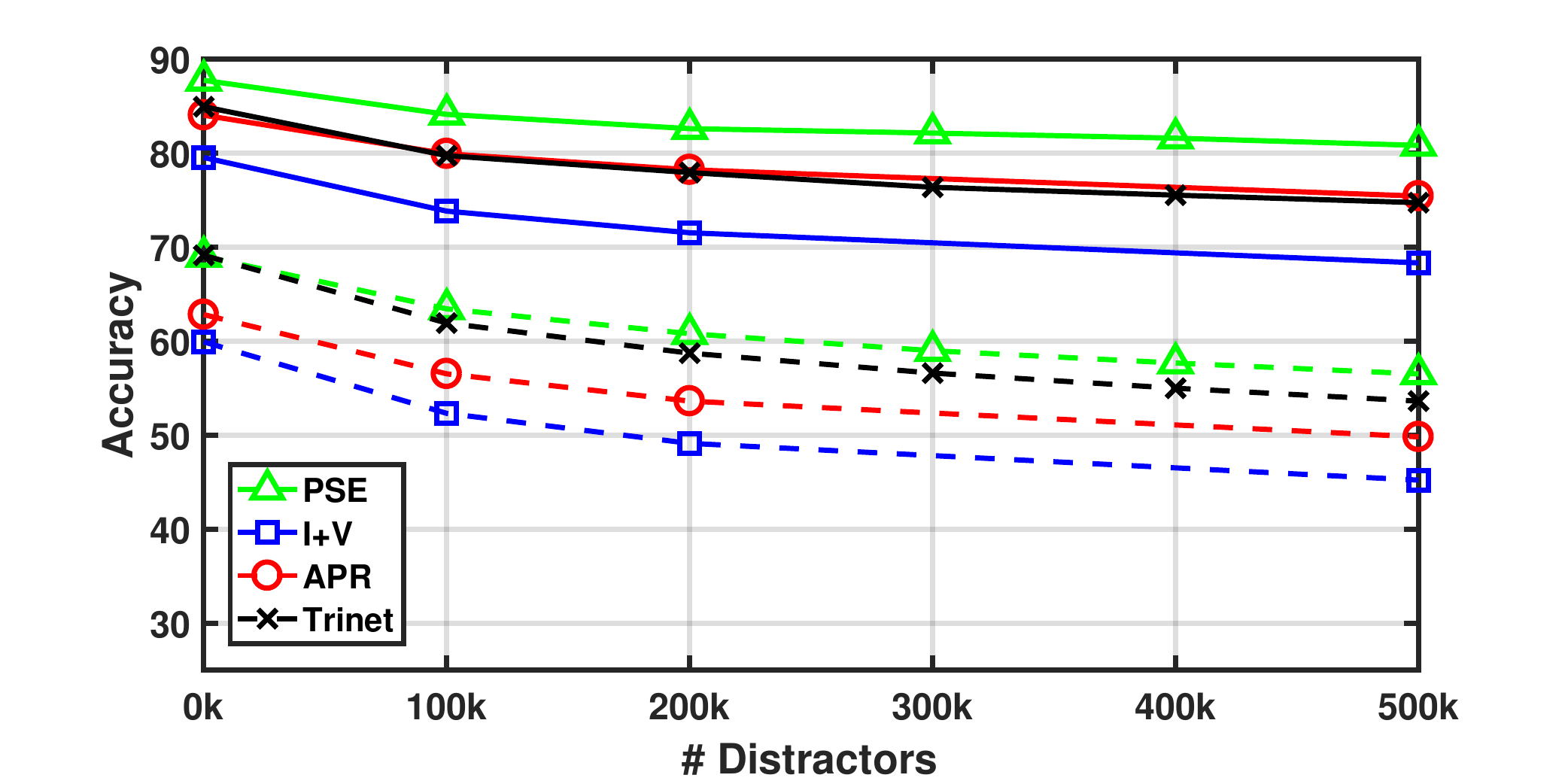}
\caption{Comparison of accuracies of the proposed PSE embedding with I+V~\cite{IV}, APR~\cite{attid} and Trinet~\cite{trinet} with increasing gallery sizes. While solid lines show rank-1 accuracies, dashed lines show mAP accuracies. The improved scalability for large gallery sizes is indicated by the less steep drop of the PSE curves compared to those of related approaches.m}
\label{figure:market500k}
\end{figure}

In Table~\ref{table:market500k} the performance of related approaches is compared to the three proposed ResNet-50 based embeddings (view model, pose model and the combined PSE model).
For this evaluation, the stated number of distractors is added to the gallery with an original size of 19,732 images.
While the views only model is slightly better in rank-1, the PSE model performs best in regards of mAP with increasing gallery sizes.

Furthermore, in Table~\ref{table:market500kRelative}, the relative decrease in detection accuracies for these embeddings is shown. 
Here, the significant difference between the proposed model and related work becomes most visible.
Whereas the best related method (TriNet \cite{trinet}) is decreasing by $22.4\%$ in mAP and $12.0\%$ in rank-1 when adding 500k distractors, the PSE approach only loses $18.1\%$ mAP and $7.9\%$ rank-1.
This indicates the positive impact of including pose information for person re-id towards large gallery sizes.
When comparing the performance drops of the view only and the pose only embeddings, it is observed that view information seems to help robustness against large gallery sizes more than pose information.

Figure~\ref{figure:market500k} further visualizes the improved scalability by a less steep drop in accuracies of PSE in comparison to related works.
This graph also indicates that the drop in accuracy is most significant with the first 100k distractors added to the gallery. 
Afterwards the drop slows down for all models.

\subsection{Working with Automated Person Detections}
\label{section:prw}

Most Person re-id systems rely on cutout person images or bounding boxes of persons.
For many scientific datasets, these are provided either as hand-drawn boxes (\eg Duke) or automatically detected boxes validated by comparing them with hand-drawn boxes (\eg Market).
In real world scenarios, however, no hand-drawn annotations are available.
Instead, an automated person detector is used to find pedestrians in video stream images which is referred to as person search.

\begin{figure}
\centering
\includegraphics[width=\columnwidth]{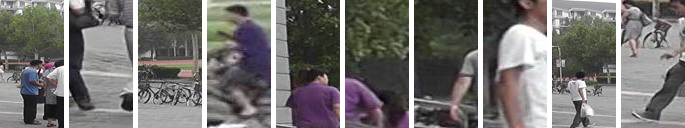}
\caption{Example images of the PRW test set showing false and badly aligned person detections.}
\label{figure:prwExamples}
\end{figure}

The Person Re-Identification in the Wild (PRW) dataset offers such automated detections created by a DPM detector alongside with a detection confidence score.
It is based on the same data source as the Market dataset and provides 13,126 training, 140,469 test and 2,057 query images cut out from video streams of six cameras.
While the videos used to create the dataset are the same as for Market, person identities are not consistent and there are only 482 IDs in the training and 450 IDs in the test set (see Table~\ref{table:prwDataset}).
Since the PRW datasets provides automatically detected person images, false detections and badly aligned images are contained.
Figure~\ref{figure:prwExamples} shows examples of such images.

\begin{table*}
  	\centering
    \begin{tabular}{ | c c | c c | c | c |}
        \hline
        \multicolumn{2}{|c|}{train}	& \multicolumn{2}{c|}{test} & query 		&  \\
        IDs 		& images		& IDs	 	& images 		& images 	& cameras \\
        \hline
        482 		& 13,126 		& 450		& 140,469		& 2,057			& 6	\\
        \hline
    \end{tabular}
    \caption{Characteristics of the PRW dataset.}
    \label{table:prwDataset}
\end{table*}

While the confidence scores indicate the confidence of a detection being a person, one still has to choose a threshold for which detections to be used.
While selecting a higher threshold keeps out most false detections and badly aligned images, it also drops out relevant person detections.
On the other hand, using all detections tends to drop person re-id accuracy as more images are in the gallery set and thus can be mixed up by the re-id system.

To evaluate a person re-id system in this context, the PRW evaluation protocol calculates mAP and rank accuracies for different detector confidence thresholds.
Since with different thresholds different numbers of persons are detected, the average number of detections per video frame is used to track re-id performance. 
The average number of actual persons per frame in the PRW dataset is about three.

Table~\ref{table:prw} compares the state of the art of the PRW dataset with results of the proposed approaches for an average number of 3, 5, 10 and 20 detections per frame.
Whereas the paper of Zheng \etal \cite{PRW} provides scores for an average number of 3, 5 and 10 detections, the paper of Jimin \etal \cite{IAN} only provides mAP and rank-1 for an average of 3 detections per frame.

\begin{table*}[b]
\centering
\resizebox{\linewidth}{!}{
\begin{tabular}{| l | l | ccc | ccc | ccc | ccc |}
    \hline
    Detector 	& Method 		& \multicolumn{3}{c|}{\#detections=3} 	& \multicolumn{3}{c|}{\#detections=5} 	& \multicolumn{3}{c|}{\#detections=10} 	& \multicolumn{3}{c|}{\#detections=20} \\
    			& & mAP & R-1 & R-20 						& mAP & R-1 & R-20 						& mAP & R-1 & R-20 						& mAP & R-1 & R-20 \\
	\hline\hline
    DPM & IDE$_{det} \cite{PRW}$			& 17.2 		& 45.9 		& 77.9 		& 18.8  	& 45.9 		& 77.4 		& 19.2  	& 45.7 		& 76.0		&&& \\
    DPM-Alex & IDE$_{det}$ \cite{PRW} 		& 20.2 		& 48.2 		& 78.1 		& 20.3 		& 47.4		& 77.1		& 19.9 		& 47.2 		& 76.4		&&& \\
    DPM-Alex & IDE$_{det}$+CWS \cite{PRW} 	& 20.0 		& 48.2 		& 78.8 		& 20.5 		& 48.3		& 78.8		& 20.5 		& 48.3 		& 78.8		&&& \\
     \multicolumn{2}{|l|}{IAN (ResNet-101) \cite{IAN}} 	& 23.0 		& 61.9		& 			& 			& 			& 			&			& 			& 			&&& \\
    \hline
    DPM & Baseline 					& 25.4		& 59.0		& 83.9		& 27.5		& 59.1		& 83.9		& 28.3		& 58.1		& 83.3		& 28.5		& 57.0		& 82.9 \\
    DPM & View only 				& 28.5		& 63.4		& 87.3		& 30.8		& 63.1		& 86.8		& 31.4		& 62.0		& 86.1		& 31.4		& 61.3		& 85.4 \\
    DPM & Pose only 				& 26.2		& 59.1		& 84.6 		& 28.4		& 58.6		& 84.4		& 29.1		& 58.1		& 83.4		& 29.4		& 57.6		& 83.0 \\
    DPM & PSE						& \f{29.3}	& \f{65.1}	& \f{88.3}	& \f{31.7}	& \f{65.0}	& \f{88.2}	& \f{32.4}	& \f{64.5}	& \f{87.5}	& \f{32.6}	& \f{63.9}	& \f{87.0} \\
	\hline
\end{tabular}
}
\caption{Comparison of the proposed approach (ResNet-50 based) with the state of the art for the PRW dataset. The different detector and model combinations are evaluated with a varying number of average person detections per frame.}
\label{table:prw}
\end{table*}

In comparison to the state of the art, all proposed approaches achieve improvements with the final PSE embedding having a $6.3\%$ higher mAP and $3.2\%$ higher rank-1 score.
Furthermore, the models show a very low reduction in rank-1 with increasing number of detections per frame, \ie with an increasing number of false detections and badly aligned images being processed.
Moreover, mAP even increases by $3.3\%$ from 3 to 20 detections for the PSE model.

When comparing the given results with the results of \cite{PRW}, large gains are observed while only the simple DPM detected boxes (instead of DPM-Alex) are used and no \emph{Confidence Weighted Similarity} (CWS) \cite{PRW} is applied.
Both strategies, using a better detector and utilizing the detection scores to weight re-id distances, are independent of the underlying re-id system and thus can also be applied to the proposed approaches which should improve results further.

Investigating the differences between the proposed three methods, it is again observed that the view information seems more beneficial than pose information as a much higher gain over the baseline model is achieved.
While mAP is improved, the rank-1 accuracy of the pose model performs even or slightly worse than the baseline model for an increasing number of detections.
Interestingly, the combined PSE model improves by almost $2\%$ in rank-1 over the views only model showing that pose information benefits re-id with this model significantly.
Furthermore, rank-1 accuracy is dropping significantly less steeply for the PSE model ($-1.2\%$) than for the views ($-2.1\%$), pose ($1.5\%$), and baseline model ($2.0\%$) while mAP is increasing more.

%% file: content/conclusion.tex

\chapter{Conclusion}
\label{ch:Conclusion}

In this work two ways of incorporating pose information into a CNN based re-id system have been presented allowing it to exploit fine-grained pose information in the form of joint locations as well as coarse pose information in the form of camera view angles.
Furthermore, due to the straightforward design of the proposed network extensions they can easily be integrated into standard CNN architectures like ResNet-50 and Inception-v4.
Moreover, it has been shown that the combined use of both pose information by combining the two extensions improves re-id further.
The combined method PSE sets a new state of the art on the four widely used challenging person re-id datasets Market, Duke, MARS and PRW.
Additionally, X-MARS, a reordering of the MARS dataset has been proposed allowing meaningful cross-evaluation of single-image trained models on tracklet data giving insights into real world considerations.

The extensive evaluation of the effects of including different degrees of pose information into the proposed models shows the significance of fine-grained and coarse pose information.
While the fine-grained joint locations offer a less consistent improvement, the incorporation of view information consistently shows significant gains over the baseline model.
Moreover, the combination of view and pose information yields even better results.
To be able to train the model's internal view prediction for datasets not offering view labels, it has been shown that pre-training the view predictor on the RAP dataset is a viable option.
Nevertheless, if actual view labels would be available, the training procedure could be simplified and performance might improve further.

Another aspect left to future works on this topic is the analysis of effects of varying numbers of view units.
For example, it could be interesting to add an additional fourth view unit which is not weighted to cover features occurring in all views or to add more view units to realize a more fine-grained view discretization.

The introduction of the X-MARS reordering of the MARS dataset allows a meaningful evaluation of image based re-id systems on tracklet data. 
It has been demonstrated that training a model on image data to use it to process tracklet data is a viable option reducing the required labeling from complete tracklets to single images.

The evaluation on the Market500k dataset shows that another strength of the pose sensitive re-id embedding is its improved scalability for very large gallery sizes.
Again, the combination of view and pose information yields an even more scalable embedding, which is dropping less in accuracies with more images being added to the gallery.

On the PRW dataset, investigation of the resistance of the PSE model against false detections and badly aligned person images showed its benefits over state of the art methods.
A further observation is that the improvement over the baseline achieved by the PSE model is larger than just the sum of the improvements of the view and the pose model alone, stressing the importance and potential of combining different pose information.

However, the evaluation shows that the pose only model does not achieve significant improvements over the baseline on the PRW dataset.
This suggests that this model cannot utilize the pose maps to detect false detections well although the pose estimator produces reasonable pose maps for this case.
One reason for this could be the fact that during training no false detections and bad aligned images are learned.
Adding a false detection class to the training set to allow the network training and detection of non-person images could improve this.
Moreover, embedding the pose estimator into the model to allow an end-to-end training could also result in better performance since it would extend the idea of feeding the unmodified pose maps to allowing the network to find a suitable representation of the pose information.

%% file: thesis.bbl
\providecommand{\etalchar}[1]{$^{#1}$}
\begin{thebibliography}{ZWW{\etalchar{+}}17}
\providecommand{\url}[1]{#1}
\csname url@samestyle\endcsname
\providecommand{\newblock}{\relax}
\providecommand{\bibinfo}[2]{#2}
\providecommand{\BIBentrySTDinterwordspacing}{\spaceskip=0pt\relax}
\providecommand{\BIBentryALTinterwordstretchfactor}{4}
\providecommand{\BIBentryALTinterwordspacing}{\spaceskip=\fontdimen2\font plus
\BIBentryALTinterwordstretchfactor\fontdimen3\font minus
  \fontdimen4\font\relax}
\providecommand{\BIBforeignlanguage}[2]{{%
\expandafter\ifx\csname l@#1\endcsname\relax
\typeout{** WARNING: IEEEtranSA.bst: No hyphenation pattern has been}%
\typeout{** loaded for the language `#1'. Using the pattern for}%
\typeout{** the default language instead.}%
\else
\language=\csname l@#1\endcsname
\fi
#2}}
\providecommand{\BIBdecl}{\relax}
\BIBdecl

\bibitem[AJM15]{ahmed2015improved}
E.~Ahmed, M.~Jones, and T.~K. Marks, ``An improved deep learning architecture
  for person re-identification,'' in \emph{Proceedings of the IEEE Conference
  on Computer Vision and Pattern Recognition}, 2015, pp. 3908--3916.

\bibitem[BBT17]{Bai_2017_CVPR}
S.~Bai, X.~Bai, and Q.~Tian, ``Scalable person re-identification on supervised
  smoothed manifold,'' in \emph{The IEEE Conference on Computer Vision and
  Pattern Recognition (CVPR)}, July 2017.

\bibitem[BEBV08]{bouwmans2008background}
T.~Bouwmans, F.~El~Baf, and B.~Vachon, ``Background modeling using mixture of
  gaussians for foreground detection-a survey,'' \emph{Recent Patents on
  Computer Science}, vol.~1, no.~3, pp. 219--237, 2008.

\bibitem[CC14]{cheng2014person}
D.~S. Cheng and M.~Cristani, ``Person re-identification by articulated
  appearance matching,'' in \emph{Person Re-Identification}.\hskip 1em plus
  0.5em minus 0.4em\relax Springer, 2014, pp. 139--160.

\bibitem[CCS{\etalchar{+}}11]{cheng2011custom}
D.~S. Cheng, M.~Cristani, M.~Stoppa, L.~Bazzani, and V.~Murino, ``Custom
  pictorial structures for re-identification.'' in \emph{BMVC}, vol.~2, no.~5,
  2011, p.~6.

\bibitem[CGZ{\etalchar{+}}16]{cheng2016person}
D.~Cheng, Y.~Gong, S.~Zhou, J.~Wang, and N.~Zheng, ``Person re-identification
  by multi-channel parts-based cnn with improved triplet loss function,'' in
  \emph{Proceedings of the IEEE Conference on Computer Vision and Pattern
  Recognition}, 2016, pp. 1335--1344.

\bibitem[CP06]{cheng2006matching}
E.~D. Cheng and M.~Piccardi, ``Matching of objects moving across disjoint
  cameras,'' in \emph{Image Processing, 2006 IEEE International Conference
  on}.\hskip 1em plus 0.5em minus 0.4em\relax IEEE, 2006, pp. 1769--1772.

\bibitem[CY16]{cho2016improving}
Y.-J. Cho and K.-J. Yoon, ``Improving person re-identification via pose-aware
  multi-shot matching,'' in \emph{Proceedings of the IEEE Conference on
  Computer Vision and Pattern Recognition}, 2016, pp. 1354--1362.

\bibitem[CZG17]{DPFL}
Y.~Chen, X.~Zhu, and S.~Gong, ``Person re-identification by deep learning
  multi-scale representations,'' in \emph{ICCV workshop on cross domain human
  identification}, 2017.

\bibitem[DSTR11]{doretto2011appearance}
G.~Doretto, T.~Sebastian, P.~Tu, and J.~Rittscher, ``Appearance-based person
  reidentification in camera networks: problem overview and current
  approaches,'' \emph{Journal of Ambient Intelligence and Humanized Computing},
  vol.~2, no.~2, pp. 127--151, 2011.

\bibitem[DWSP12]{dollar2012pedestrian}
P.~Dollar, C.~Wojek, B.~Schiele, and P.~Perona, ``Pedestrian detection: An
  evaluation of the state of the art,'' \emph{IEEE transactions on pattern
  analysis and machine intelligence}, vol.~34, no.~4, pp. 743--761, 2012.

\bibitem[EF10]{eichner2010we}
M.~Eichner and V.~Ferrari, ``We are family: Joint pose estimation of multiple
  persons,'' in \emph{European conference on computer vision}.\hskip 1em plus
  0.5em minus 0.4em\relax Springer, 2010, pp. 228--242.

\bibitem[FBP{\etalchar{+}}10]{farenzena2010person}
M.~Farenzena, L.~Bazzani, A.~Perina, V.~Murino, and M.~Cristani, ``Person
  re-identification by symmetry-driven accumulation of local features,'' in
  \emph{Computer Vision and Pattern Recognition (CVPR), 2010 IEEE Conference
  on}.\hskip 1em plus 0.5em minus 0.4em\relax IEEE, 2010, pp. 2360--2367.

\bibitem[GT08]{gray2008viewpoint}
D.~Gray and H.~Tao, ``Viewpoint invariant pedestrian recognition with an
  ensemble of localized features,'' in \emph{European conference on computer
  vision}.\hskip 1em plus 0.5em minus 0.4em\relax Springer, 2008, pp. 262--275.

\bibitem[HBL17]{trinet}
A.~Hermans, L.~Beyer, and B.~Leibe, ``In defense of the triplet loss for person
  re-identification,'' \emph{arXiv preprint arXiv:1703.07737}, 2017.

\bibitem[HZRS16]{ResNet}
K.~He, X.~Zhang, S.~Ren, and J.~Sun, ``Deep residual learning for image
  recognition,'' in \emph{Proceedings of the IEEE conference on computer vision
  and pattern recognition}, 2016, pp. 770--778.

\bibitem[IPA{\etalchar{+}}]{DeeperCut}
E.~Insafutdinov, L.~Pishchulin, B.~Andres, M.~Andriluka, and B.~Schiele.

\bibitem[JM08]{jiang2008global}
H.~Jiang and D.~R. Martin, ``Global pose estimation using non-tree models,'' in
  \emph{Computer Vision and Pattern Recognition, 2008. CVPR 2008. IEEE
  Conference on}.\hskip 1em plus 0.5em minus 0.4em\relax IEEE, 2008, pp. 1--8.

\bibitem[KHW{\etalchar{+}}12]{koestinger2012large}
M.~Koestinger, M.~Hirzer, P.~Wohlhart, P.~M. Roth, and H.~Bischof, ``Large
  scale metric learning from equivalence constraints,'' in \emph{Computer
  Vision and Pattern Recognition (CVPR), 2012 IEEE Conference on}.\hskip 1em
  plus 0.5em minus 0.4em\relax IEEE, 2012, pp. 2288--2295.

\bibitem[KSH12]{krizhevsky2012imagenet}
A.~Krizhevsky, I.~Sutskever, and G.~E. Hinton, ``Imagenet classification with
  deep convolutional neural networks,'' in \emph{Advances in neural information
  processing systems}, 2012, pp. 1097--1105.

\bibitem[LB{\etalchar{+}}95]{lecun1995convolutional}
Y.~LeCun, Y.~Bengio \emph{et~al.}, ``Convolutional networks for images, speech,
  and time series,'' \emph{The handbook of brain theory and neural networks},
  vol. 3361, no.~10, p. 1995, 1995.

\bibitem[LCZH17]{Li_2017_CVPR}
D.~Li, X.~Chen, Z.~Zhang, and K.~Huang, ``Learning deep context-aware features
  over body and latent parts for person re-identification,'' in \emph{The IEEE
  Conference on Computer Vision and Pattern Recognition (CVPR)}, July 2017.

\bibitem[LHGM12]{layne2012person}
R.~Layne, T.~M. Hospedales, S.~Gong, and Q.~Mary, ``Person re-identification by
  attributes.'' in \emph{Bmvc}, vol.~2, no.~3, 2012, p.~8.

\bibitem[LRL{\etalchar{+}}17]{ConsAw}
J.~Lin, L.~Ren, J.~Lu, J.~Feng, and J.~Zhou, ``Consistent-aware deep learning
  for person re-identification in a camera network,'' in \emph{The IEEE
  Conference on Computer Vision and Pattern Recognition (CVPR)}, July 2017.

\bibitem[LTZ13]{ladicky2013human}
L.~Ladicky, P.~H. Torr, and A.~Zisserman, ``Human pose estimation using a joint
  pixel-wise and part-wise formulation,'' in \emph{proceedings of the IEEE
  Conference on Computer Vision and Pattern Recognition}, 2013, pp. 3578--3585.

\bibitem[LYO17]{QMA}
Y.~Liu, J.~Yan, and W.~Ouyang, ``Quality aware network for set to set
  recognition,'' in \emph{The IEEE Conference on Computer Vision and Pattern
  Recognition (CVPR)}, 2017.

\bibitem[LZC{\etalchar{+}}16]{RapDataset}
D.~Li, Z.~Zhang, X.~Chen, H.~Ling, and K.~Huang, ``A richly annotated dataset
  for pedestrian attribute recognition,'' \emph{arXiv preprint
  arXiv:1603.07054}, 2016.

\bibitem[LZG17]{JLML}
W.~Li, X.~Zhu, and S.~Gong, ``Person re-identification by deep joint learning
  of multi-loss classification,'' in \emph{International Joint Conference of
  Artificial Intelligence}, 2017.

\bibitem[LZXW14]{li2014deepreid}
W.~Li, R.~Zhao, T.~Xiao, and X.~Wang, ``Deepreid: Deep filter pairing neural
  network for person re-identification,'' in \emph{Proceedings of the IEEE
  Conference on Computer Vision and Pattern Recognition}, 2014, pp. 152--159.

\bibitem[LZZ{\etalchar{+}}17]{attid}
Y.~Lin, L.~Zheng, Z.~Zheng, Y.~Wu, and Y.~Yang, ``Improving person
  re-identification by attribute and identity learning,'' \emph{arXiv preprint
  arXiv:1703.07220}, 2017.

\bibitem[MB06]{man2006individual}
J.~Man and B.~Bhanu, ``Individual recognition using gait energy image,''
  \emph{IEEE transactions on pattern analysis and machine intelligence},
  vol.~28, no.~2, pp. 316--322, 2006.

\bibitem[PIT{\etalchar{+}}16]{deepCut}
L.~Pishchulin, E.~Insafutdinov, S.~Tang, B.~Andres, M.~Andriluka, P.~V. Gehler,
  and B.~Schiele, ``Deepcut: Joint subset partition and labeling for multi
  person pose,'' in \emph{Proceedings of the IEEE Conference on Computer Vision
  and Pattern Recognition}, 2016, pp. 4929--4937.

\bibitem[RKT75]{roebuck1975engineering}
J.~A. Roebuck, K.~H. Kroemer, and W.~G. Thomson, \emph{Engineering
  anthropometry methods}.\hskip 1em plus 0.5em minus 0.4em\relax John Wiley \&
  Sons, 1975, vol.~3.

\bibitem[RLT{\etalchar{+}}17]{rahimpour2017person}
A.~Rahimpour, L.~Liu, A.~Taalimi, Y.~Song, and H.~Qi, ``Person
  re-identification using visual attention,'' \emph{arXiv preprint
  arXiv:1707.07336}, 2017.

\bibitem[RSZ{\etalchar{+}}16a]{DukeDataset}
E.~Ristani, F.~Solera, R.~Zou, R.~Cucchiara, and C.~Tomasi, ``Performance
  measures and a data set for multi-target, multi-camera tracking,'' in
  \emph{European Conference on Computer Vision workshop on Benchmarking
  Multi-Target Tracking}, 2016.

\bibitem[RSZ{\etalchar{+}}16b]{DukeMTMC}
E.~Ristani, F.~Solera, R.~Zou, R.~Cucchiara, and C.~Tomasi, ``Performance
  measures and a data set for multi-target, multi-camera tracking,'' in
  \emph{European Conference on Computer Vision workshop on Benchmarking
  Multi-Target Tracking}, 2016.

\bibitem[Sat13]{satta2013appearance}
R.~Satta, ``Appearance descriptors for person re-identification: a
  comprehensive review,'' \emph{arXiv preprint arXiv:1307.5748}, 2013.

\bibitem[SIVA17]{InceptionV4}
C.~Szegedy, S.~Ioffe, V.~Vanhoucke, and A.~A. Alemi, ``Inception-v4,
  inception-resnet and the impact of residual connections on learning.'' in
  \emph{AAAI}, vol.~4, 2017, p.~12.

\bibitem[SLZ{\etalchar{+}}17]{su2017pose}
C.~Su, J.~Li, S.~Zhang, J.~Xing, W.~Gao, and Q.~Tian, ``Pose-driven deep
  convolutional model for person re-identification,'' in \emph{Proceedings of
  the IEEE Conference on Computer Vision ICCV}, 2017, pp. 3960--3969.

\bibitem[SNV99]{stevenage1999visual}
S.~V. Stevenage, M.~S. Nixon, and K.~Vince, ``Visual analysis of gait as a cue
  to identity,'' \emph{Applied cognitive psychology}, vol.~13, no.~6, pp.
  513--526, 1999.

\bibitem[Spr16]{MarsDataset}
\emph{MARS: A Video Benchmark for Large-Scale Person Re-identification}.\hskip
  1em plus 0.5em minus 0.4em\relax Springer, 2016.

\bibitem[SS17]{ACRN}
A.~Schumann and R.~Stiefelhagen, ``Person re-identification by deep learning
  attribute-complementary information,'' in \emph{Computer Vision and Pattern
  Recognition Workshops (CVPRW), 2017 IEEE Conference on}.\hskip 1em plus 0.5em
  minus 0.4em\relax IEEE, 2017, pp. 1435--1443.

\bibitem[SSES17]{pse-ecn}
\BIBentryALTinterwordspacing
M.~S. Sarfraz, A.~Schumann, A.~Eberle, and R.~Stiefelhagen, ``A pose-sensitive
  embedding for person re-identification with expanded cross neighborhood
  re-ranking,'' \emph{CoRR}, vol. abs/1711.10378, 2017. [Online]. Available:
  \url{http://arxiv.org/abs/1711.10378}
\BIBentrySTDinterwordspacing

\bibitem[SSWS17]{SarfrazPedestrian17}
\BIBentryALTinterwordspacing
M.~S. Sarfraz, A.~Schumann, Y.~Wang, and R.~Stiefelhagen, ``Deep view-sensitive
  pedestrian attribute inference in an end-to-end model,'' \emph{CoRR}, vol.
  abs/1707.06089, 2017. [Online]. Available:
  \url{http://arxiv.org/abs/1707.06089}
\BIBentrySTDinterwordspacing

\bibitem[SZDW17]{SVD}
Y.~Sun, L.~Zheng, W.~Deng, and S.~Wang, ``Svdnet for pedestrian retrieval,'' in
  \emph{The IEEE International Conference on Computer Vision (ICCV)}, Oct 2017.

\bibitem[VSL{\etalchar{+}}16]{varior2016siamese}
R.~R. Varior, B.~Shuai, J.~Lu, D.~Xu, and G.~Wang, ``A siamese long short-term
  memory architecture for human re-identification,'' in \emph{European
  Conference on Computer Vision}.\hskip 1em plus 0.5em minus 0.4em\relax
  Springer, 2016, pp. 135--153.

\bibitem[WDS{\etalchar{+}}07]{wang2007shape}
X.~Wang, G.~Doretto, T.~Sebastian, J.~Rittscher, and P.~Tu, ``Shape and
  appearance context modeling,'' in \emph{Computer Vision, 2007. ICCV 2007.
  IEEE 11th International Conference on}.\hskip 1em plus 0.5em minus
  0.4em\relax IEEE, 2007, pp. 1--8.

\bibitem[WSH16]{wu2016personnet}
L.~Wu, C.~Shen, and A.~v.~d. Hengel, ``Personnet: Person re-identification with
  deep convolutional neural networks,'' \emph{arXiv preprint arXiv:1601.07255},
  2016.

\bibitem[XJRN03]{xing2003distance}
E.~P. Xing, M.~I. Jordan, S.~J. Russell, and A.~Y. Ng, ``Distance metric
  learning with application to clustering with side-information,'' in
  \emph{Advances in neural information processing systems}, 2003, pp. 521--528.

\bibitem[XLW{\etalchar{+}}17]{OIM}
T.~Xiao, S.~Li, B.~Wang, L.~Lin, and X.~Wang, ``Joint detection and
  identification feature learning for person search,'' in \emph{Proc. CVPR},
  2017.

\bibitem[XXT{\etalchar{+}}17]{IAN}
\BIBentryALTinterwordspacing
J.~Xiao, Y.~Xie, T.~Tillo, K.~Huang, Y.~Wei, and J.~Feng, ``{IAN:} the
  individual aggregation network for person search,'' \emph{CoRR}, vol.
  abs/1705.05552, 2017. [Online]. Available:
  \url{http://arxiv.org/abs/1705.05552}
\BIBentrySTDinterwordspacing

\bibitem[YJ06]{yang2006distance}
L.~Yang and R.~Jin, ``Distance metric learning: A comprehensive survey,''
  \emph{Michigan State Universiy}, vol.~2, no.~2, 2006.

\bibitem[YLLL14]{yi2014deep}
D.~Yi, Z.~Lei, S.~Liao, and S.~Z. Li, ``Deep metric learning for person
  re-identification,'' in \emph{Pattern Recognition (ICPR), 2014 22nd
  International Conference on}.\hskip 1em plus 0.5em minus 0.4em\relax IEEE,
  2014, pp. 34--39.

\bibitem[YMZ{\etalchar{+}}17]{DGM}
M.~Ye, A.~J. Ma, L.~Zheng, J.~Li, and P.~C. Yuen, ``Dynamic label graph
  matching for unsupervised video re-identification,'' in \emph{The IEEE
  International Conference on Computer Vision (ICCV)}, Oct 2017.

\bibitem[YZBB17]{yu2017divide}
R.~Yu, Z.~Zhou, S.~Bai, and X.~Bai, ``Divide and fuse: A re-ranking approach
  for person re-identification,'' in \emph{BMVC}, 2017.

\bibitem[ZHLY17]{poseInvariantEmbedding}
\BIBentryALTinterwordspacing
L.~Zheng, Y.~Huang, H.~Lu, and Y.~Yang, ``Pose invariant embedding for deep
  person re-identification,'' \emph{CoRR}, vol. abs/1701.07732, 2017. [Online].
  Available: \url{http://arxiv.org/abs/1701.07732}
\BIBentrySTDinterwordspacing

\bibitem[ZHW{\etalchar{+}}17]{Forest}
Z.~Zhou, Y.~Huang, W.~Wang, L.~Wang, and T.~Tan, ``See the forest for the
  trees: Joint spatial and temporal recurrent neural networks for video-based
  person re-identification,'' in \emph{The IEEE Conference on Computer Vision
  and Pattern Recognition (CVPR)}, July 2017.

\bibitem[ZLWZ17]{zhao2017deeply}
L.~Zhao, X.~Li, J.~Wang, and Y.~Zhuang, ``Deeply-learned part-aligned
  representations for person re-identification,'' \emph{ICCV}, 2017.

\bibitem[ZST{\etalchar{+}}15]{Market1501Dataset}
L.~Zheng, L.~Shen, L.~Tian, S.~Wang, J.~Wang, and Q.~Tian, ``Scalable person
  re-identification: A benchmark,'' in \emph{Computer Vision, IEEE
  International Conference on}, 2015.

\bibitem[ZTS{\etalchar{+}}17]{zhao2017spindle}
H.~Zhao, M.~Tian, S.~Sun, J.~Shao, J.~Yan, S.~Yi, X.~Wang, and X.~Tang,
  ``Spindle net: Person re-identification with human body region guided feature
  decomposition and fusion,'' in \emph{Proceedings of the IEEE Conference on
  Computer Vision and Pattern Recognition}, 2017, pp. 1077--1085.

\bibitem[ZWW{\etalchar{+}}17]{P2S}
S.~Zhou, J.~Wang, J.~Wang, Y.~Gong, and N.~Zheng, ``Point to set similarity
  based deep feature learning for person re-identification,'' in \emph{The IEEE
  Conference on Computer Vision and Pattern Recognition (CVPR)}, July 2017.

\bibitem[ZZCL17]{zhong2017re}
Z.~Zhong, L.~Zheng, D.~Cao, and S.~Li, ``Re-ranking person re-identification
  with k-reciprocal encoding,'' pp. 1318--1327, 2017.

\bibitem[ZZK05]{zajdel2005keeping}
W.~Zajdel, Z.~Zivkovic, and B.~Krose, ``Keeping track of humans: Have i seen
  this person before?'' in \emph{Robotics and Automation, 2005. ICRA 2005.
  Proceedings of the 2005 IEEE International Conference on}.\hskip 1em plus
  0.5em minus 0.4em\relax IEEE, 2005, pp. 2081--2086.

\bibitem[ZZS{\etalchar{+}}16]{PRW}
\BIBentryALTinterwordspacing
L.~Zheng, H.~Zhang, S.~Sun, M.~Chandraker, and Q.~Tian, ``Person
  re-identification in the wild,'' \emph{CoRR}, vol. abs/1604.02531, 2016.
  [Online]. Available: \url{http://arxiv.org/abs/1604.02531}
\BIBentrySTDinterwordspacing

\bibitem[ZZY16]{IV}
Z.~Zheng, L.~Zheng, and Y.~Yang, ``A discriminatively learned cnn embedding for
  person re-identification,'' \emph{arXiv preprint arXiv:1611.05666}, 2016.

\bibitem[ZZY17]{GAN}
Z.~Zheng, L.~Zheng, and Y.~Yang, ``Unlabeled samples generated by gan improve
  the person re-identification baseline in vitro,'' in \emph{The IEEE
  International Conference on Computer Vision (ICCV)}, Oct 2017.

\end{thebibliography}
